\documentclass[10pt,twocolumn,letterpaper]{article}

\usepackage{cvpr}
\usepackage{times}
\usepackage{epsfig}
\usepackage{graphicx}
\usepackage{amsmath}
\usepackage{amssymb}
\usepackage{booktabs}
\usepackage{siunitx}
\usepackage{graphicx}
\usepackage{caption}
\usepackage{subcaption}

\usepackage[pagebackref=true,breaklinks=true,letterpaper=true,colorlinks,bookmarks=false]{hyperref}

\cvprfinalcopy 
\def\cvprPaperID{xx} 

\begin{document}

\title{Deflating Dataset Bias Using Synthetic Data Augmentation}
\author{Nikita Jaipuria, Xianling Zhang, Rohan Bhasin, Mayar Arafa \\
Punarjay Chakravarty, Shubham Shrivastava, Sagar Manglani, Vidya N. Murali\\
Ford Greenfield Labs, Palo Alto\\
{\tt\footnotesize \{njaipuri, xzhan258, rbhasin, marafa, pchakra5, sshriva5, smanglan, vnariyam\}@ford.com}
}

\maketitle

\begin{abstract}
Deep Learning has seen an unprecedented increase in vision applications since the publication of large-scale object recognition datasets and introduction of scalable compute hardware. State-of-the-art methods for most vision tasks for Autonomous Vehicles (AVs) rely on supervised learning and often fail to generalize to domain shifts and/or outliers. Dataset diversity is thus key to successful real-world deployment. No matter how big the size of the dataset, capturing long tails of the distribution pertaining to task-specific environmental factors is impractical. The goal of this paper is to investigate the use of targeted synthetic data augmentation - combining the benefits of gaming engine simulations and sim2real style transfer techniques - for filling gaps in real datasets for vision tasks. Empirical studies on three different computer vision tasks of practical use to AVs - parking slot detection, lane detection and monocular depth estimation - consistently show that having synthetic data in the training mix provides a significant boost in cross-dataset generalization performance as compared to training on real data only, for the same size of the training set.
\end{abstract}

\section{Introduction}
\label{sec:intro}
Data-hungry Deep Neural Networks (DNNs) thrive when trained on large datasets. The release of large-scale datasets (such as ImageNet \cite{deng2009imagenet}, COCO \cite{lin2014microsoft}, KITTI \cite{geiger2012we} and the relatively recent BDD100K \cite{yu2018bdd100k}) coupled with progress in scalable compute has led to the use of DNNs for a wide variety of vision tasks for autonomous driving. State-of-the-art methods for most of these tasks, such as object detection, semantic segmentation and depth estimation to name a few \cite{he2017mask, pan2018spatial, godard2017unsupervised}, rely on supervised learning and often fail to generalize to unseen scenarios and/or datasets. Thus, dataset diversity is key to achieving successful deployment of DNNs for real-world vision tasks, especially in safety-critical applications.

Presence of bias in static datasets, such as selection bias, capture bias, label bias and negative set bias~\cite{torralba2011unbiased,ren2018learning} is a known problem in computer vision famously shown by the \emph{Name That Dataset} experiment from Torralba~\etal \cite{torralba2011unbiased}.
However, most of these well studied biases are task-agnostic and too general in nature.
For instance, consider the task of lane detection which is one of the most common vision applications in autonomous driving. One way of addressing generic dataset selection biases is to simply augment data from multiple sources like highways, cities etc. But no matter how big the size of the dataset, it is extremely difficult to capture long tails of the distribution, and on the contrary, as shown in~\cite{torralba2011unbiased,khosla2012undoing}, mixing different datasets often ends up hurting the final performance!
This begs the question if it is ever possible to completely avoid such biases in realistic settings by means of careful data collection~\cite{recht2019imagenet}.

\begin{figure}[t]
\centering
    \includegraphics[width=0.4\linewidth]{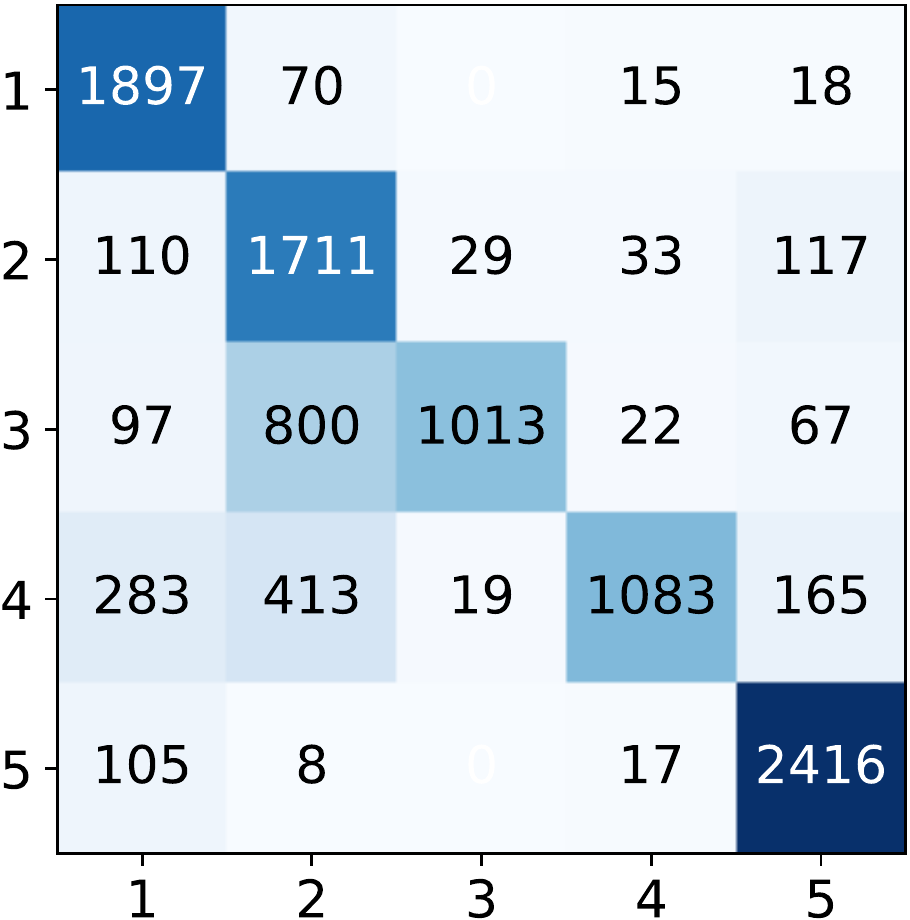}
    \includegraphics[width=0.52\linewidth]{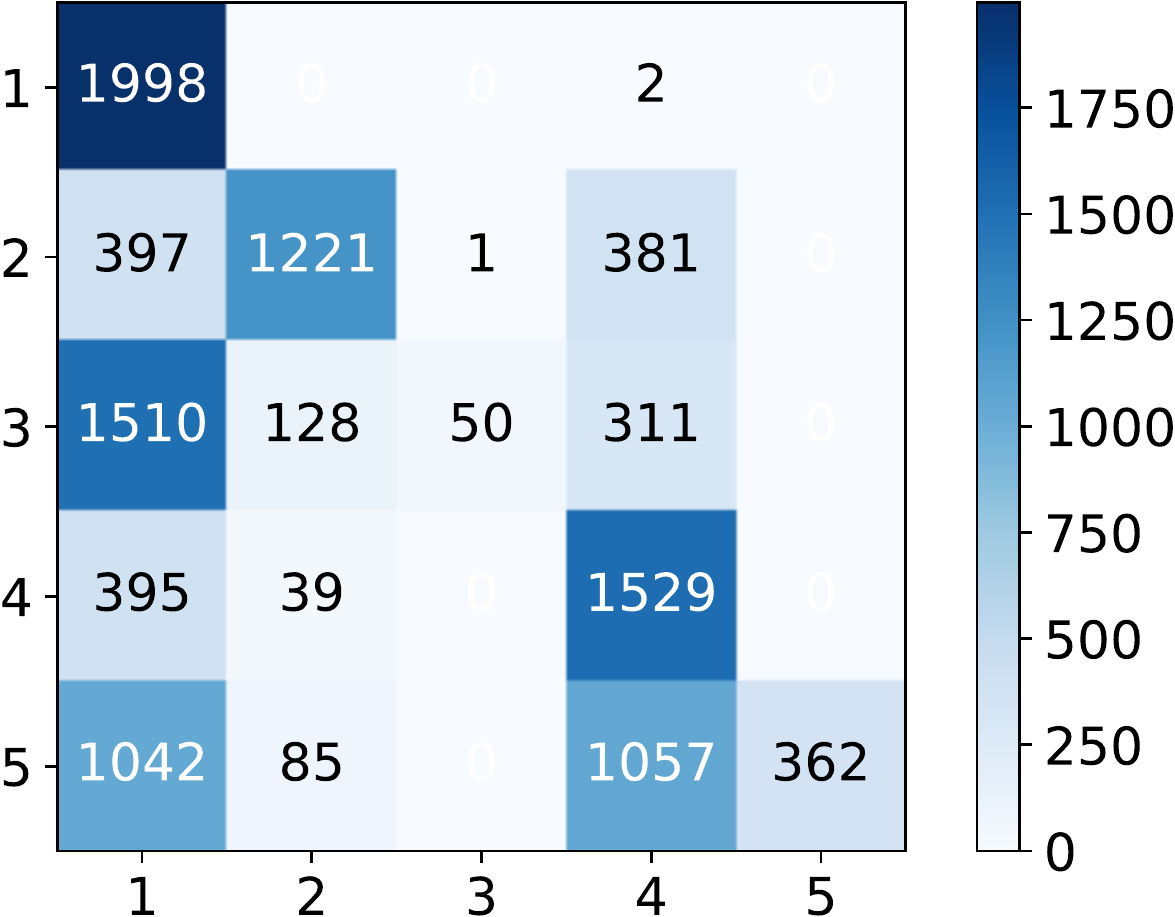}
\caption{Comparison of confusion matrices from the ResNet-50 \cite{he2016deep} based \emph{Name That Dataset} classifiers described in Section~\ref{sec:confusion} trained to distinguish between five different lane-detection datasets (left) and between the same five datasets with two of them (3 and 5) augmented with synthetic data (right). Note that synthetic data augmentation helps diffuse the strength of the diagonal indicating deflated dataset bias.\label{fig:confusion}}
\vspace{-.3in}
\end{figure}

In this work, we focus on bias in the context of the noise distribution pertaining to task-specific environmental factors. We refer to it as \emph{noise factor distribution bias}. For instance, instead of handling diversity by blindly collecting more data in our lane detection example, we chose to augment data with respect to task-specific noise factors, such as diversifying lane marker types, number of lanes in the scene, condition of lane markers, type of lane markers, weather and lighting effects etc. We show how this could go a long-way in improving algorithm performance. Hoping to obtain such targeted diversity in real data from dashboard cameras in cars is likely futile because of the time it will take and the unavailability of sources. 

One approach is to leverage advances in generative modeling to generate synthetic data for augmentation. Generative Adversarial Networks (GANs) \cite{goodfellow2014generative} have shown immense progress in the past few years in image generation \cite{karras2017progressive,karras2019style}. While they have had huge success in graphics applications \cite{park2019semantic, shaham2019singan, ledig2017photo}, synthetic data augmentation for improving performance of recognition models has seen limited success. One reason is the presence of noisy artifacts and semantic inconsistencies in the generated images \cite{isola2017image, ravuri2019classification}. Alternatively, gaming-engine simulations can be used to generate semantically consistent data of desired task-specific scenarios, but the perceptual quality is far from realistic.Why not have the best of both worlds? In contrast to performing augmentation with either generated or simulated data, we first simply simulate candidate examples and then translate via unsupervised sim2real generative models \cite{liu2017unsupervised, jaipuria2020role, zhu2017unpaired}. 

We show that this simple two-stage augmentation when targeted to encourage task-specific noise diversity leads to huge gains in cross-dataset generalization performance. We demonstrate this empirically using three different case studies of computer vision tasks in an AV perception stack: (i) parking slot detection; (ii) lane detection; and (iii) monocular depth estimation. To isolate the effect of simply training on more data, in all of these tasks, synthetic data was used to replace some amount of real data in the training set. Results showed a significant boost in cross-dataset generalization performance, especially in cases where the real dataset was small in size and heavily biased. Moreover, model performance on the original test set was not hurt which further confirms that targeted synthetic data augmentation can go a long way in enriching the real biased dataset.

\section{Related Work}
Related work on dealing with dataset bias falls under two main categories: (i) Domain Adaptation (DA); and (ii) Transfer Learning. DA is one way of dealing with inherent bias in datasets and the problem of perception algorithms failing to generalize to different datasets. Fernando \etal \cite{fernando2013unsupervised} addressed DA by learning a mapping between the source and target datasets in the form of a common subspace between their distributions. One can also learn data specific embeddings subject to minimization of MMD between them \cite{rozantsev2018beyond} in an effort to bring the two distributions closer. A classifier can then act on the learnt embeddings. Optimal transport techniques have also been used to solve DA, with \cite{bhushan2018deepjdot} minimizing the Wasserstein distance between the joint embedding and classifier label distributions of the two datasets. Wang \etal \cite{wang2018deep} provide a good taxonomy of DA techniques, including the more recent adversarial techniques based on GANs. Instead of relying on a hand-engineered loss function to bring the source and target data distributions close, these techniques use an adversarially trained discriminator network that attempts to differentiate between data from the two distributions. This discrimination can happen in: (i) the pixel space - where data from one domain is translated into the other using style transfer before being passed to the discriminator \cite{liu2016coupled, sankaranarayanan2017unsupervised}; (ii) latent space - where a discriminator learns to differentiate between the learned embeddings from the two domains \cite{sankaranarayanan2018learning} and; (ii) both the pixel and embedding space \cite{hoffman2017cycada}. In cases where one has access to unpaired and unannotated data only from the two domains, one can use cycle consistency losses \cite{liu2017unsupervised, yi2017dualgan, zhu2017unpaired} for learning a common embedding between the two spaces. Often, we are concerned with DA for a particular task - for example image segmentation or depth estimation. Recent work has shown that using losses from an auxiliary task like image segmentation can help regularize the feature embeddings \cite{hoffman2017cycada, sankaranarayanan2018learning}. These methods are most relevant to our work and future work will investigate how they fare against our approach of targeted synthetic data augmentation.

Transfer Learning is another way of dealing with dataset bias \cite{sung2018learning}. In contrast to such approaches, our method assumes no training data is available from the target domain (both for the task network and sim2real models), and that the target task is the same as the source task. Recent work \cite{alhaija2018augmented,kar2019meta} has also focused on using synthetic data to augment real datasets for AV perception tasks. Meta-sim \cite{kar2019meta} parameterizes scene-grammar to generate a synthetic data distribution that is similar to real data and is optimized for a down-stream task and Alhaija \etal \cite{alhaija2018augmented} augment real scene backgrounds with synthetically inserted objects for improved instance segmentation and object detection performance on real datasets. Our method, in contrast, investigates a general purpose, task agnostic approach to enriching real-world datasets using synthetic data.
\section{Deflating Dataset Bias}
\label{sec:method}
The main objective of this paper is to test the hypothesis that targeted synthetic data augmentation can help deflate inherent bias in large-scale image datasets. For brevity, we will refer to this hypothesis as \texttt{H}. One way of testing \texttt{H} is to compare cross-dataset generalization performance of models trained on the original dataset (real) with models trained on augmented datasets ($\text{real} + \text{synthetic}$). In this paper, three supervised learning-based computer vision tasks: (i) parking slot detection; (ii) traffic lane detection; and (iii) monocular depth estimation are used as test-beds for the motivating hypothesis \texttt{H}, using the following methodology:
\begin{enumerate}
\itemsep0em
\item Simulate images and corresponding annotation using gaming engines for a diverse set of task-specific noise factors.
\item Use unsupervised generative modeling based sim2real methods such as \cite{liu2017unsupervised, jaipuria2020role, zhu2017unpaired} to translate the simulated images into photorealistic ones, that look like they are from the training domain.
\item Train task networks with different ratios of real and simulated data (from Step 1) or real and sim2real data (from Step 2). The size of the training set is kept constant across all experiments to isolate the improvement one can obtain by simply training on more data from the improvement due to deflated dataset bias. Also, the ratio of synthetic data in the training set was increased from 0\% to 100\% in continuous intervals of 10\%.
\item Evaluate and compare cross-dataset generalization performance of all models from Step 3.
\end{enumerate}
Sections~\ref{sec:slot},~\ref{sec:lane} and~\ref{sec:depth} describe the task-specific datasets, experiments and results.

\subsection{Revisiting ``Name That Dataset''}
\label{sec:confusion}
Torralba~\etal \cite{torralba2011unbiased} investigated the then state of object recognition datasets using the \emph{Name That Dataset} experiment in which a 12-way linear SVM classifier was trained to distinguish between 12 datasets. The results showed strong \emph{signatures} for each dataset - indicating inherent \emph{bias} - despite the best efforts of their creators. We repeat the \emph{Name That Dataset} experiment in the era of deep learning with a ResNet-50 \cite{he2016deep} (pre-trained on ImageNet) trained to distinguish between five different lane-detection datasets - ApolloScape \cite{huang2018apolloscape}, BDD100K \cite{yu2018bdd100k}, CULane \cite{pan2018spatial}, Mapillary \cite{neuhold2017mapillary} and TuSimple\footnote{\url{https://github.com/TuSimple/tusimple-benchmark/tree/master/doc/lane_detection}}. $6000$ images were randomly selected from each dataset and divided into training, validation and test sets. In a subsequent experiment, we replace $50\%$ of the real data in two datasets - CULane and TuSimple - with sim2real translated images from VAE-GAN models based off of \cite{liu2017unsupervised, jaipuria2020role} and trained on unpaired simulated and real CULane and simulated and real TuSimple images respectively. We chose to apply data augmentation to only these two datasets as they are also used for the lane detection experiments in Section~\ref{sec:lane} with readily available sim2real data on hand. Fig.~\ref{fig:confusion} compares the confusion matrices of the two classifiers, with and without synthetic data augmentation. Here, the labels 1, 2, 3, 4 and 5 denote the ApolloScape, BDD100K, CULane, Mapillary and TuSimple datasets respectively. Consistent with the motivating hypothesis \texttt{H}, synthetic data augmentation diffuses the strength of the diagonal indicating deflated dataset bias.
\section{Case Study: Parking Slot Detection}
\label{sec:slot}
\begin{figure}[h]
\centering
\includegraphics[width=0.85\linewidth]{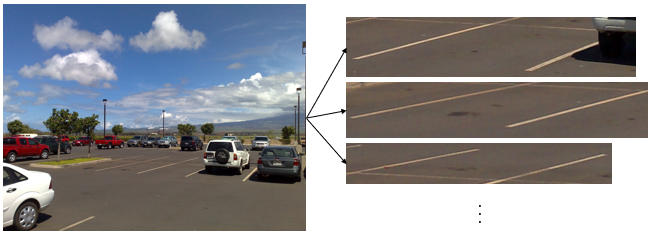}
\caption{Illustrative example of empty parking slots detected (right) in a parking lot image (left).\label{fig:slot_detection}}
\vspace{-.2in}
\end{figure}

The objective of this task is to detect empty parking slots in images taken from side vehicle cameras (see Fig.~\ref{fig:slot_detection}).
\subsection{Dataset Description}
\label{sec:data_slot}
\begin{figure}[h]
\centering
    \begin{subfigure}{.45\linewidth}
        \centering
        \includegraphics[height=.9in]{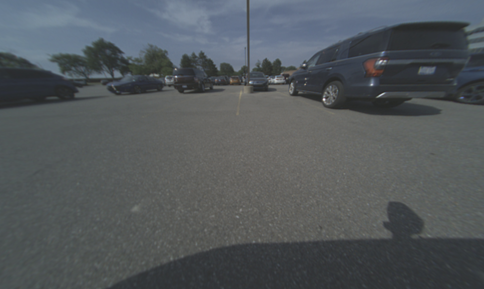}
        \caption{\small Real - Parking A}
        \label{fig:dearborn_slot}
    \end{subfigure}
    \begin{subfigure}{.45\linewidth}
        \centering
        \includegraphics[height=.9in]{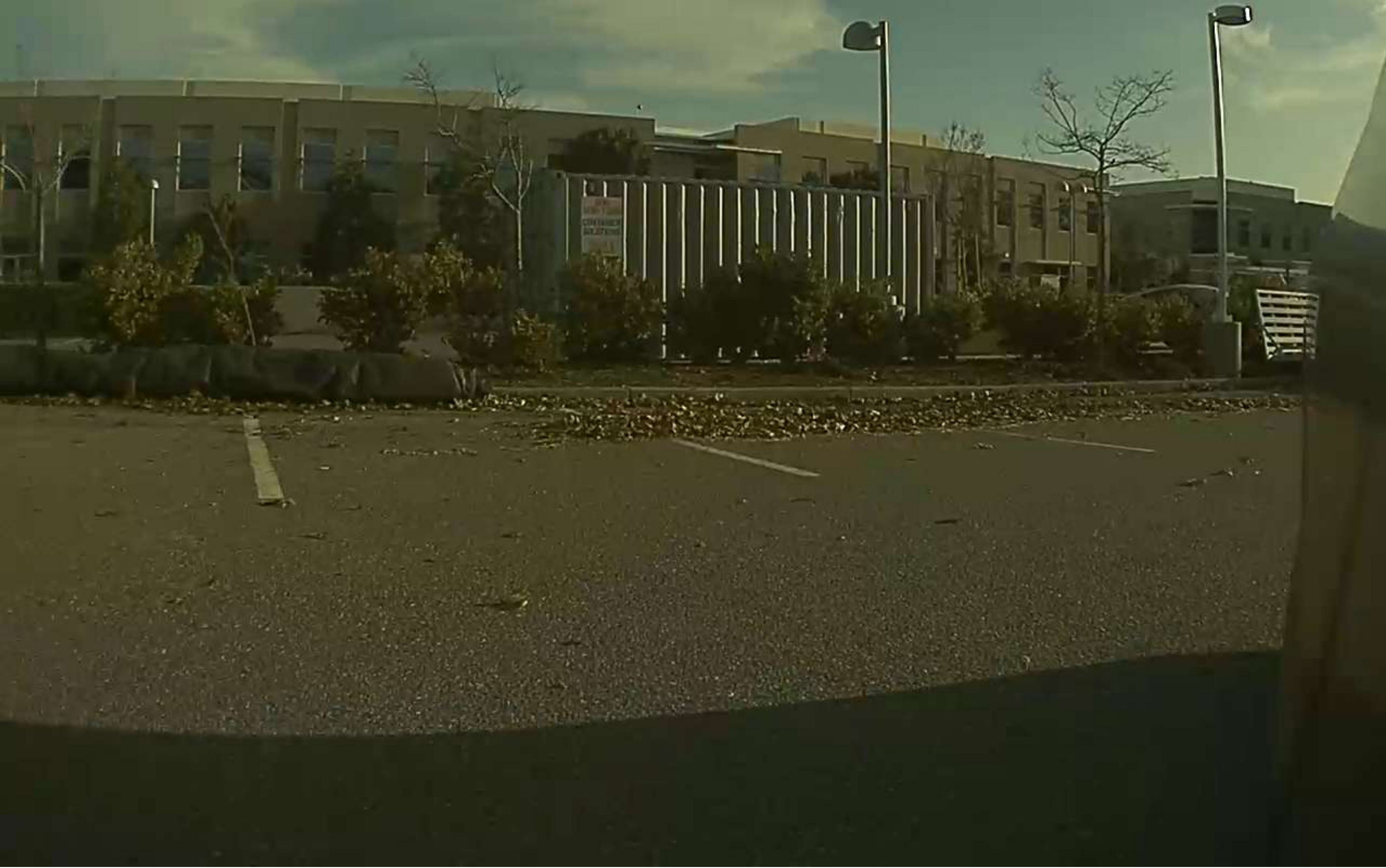}
        \caption{\small Real - Parking B}
        \label{fig:paloalto_slot}
    \end{subfigure}
    \begin{subfigure}{.45\linewidth}
        \centering
        \includegraphics[height=.925in]{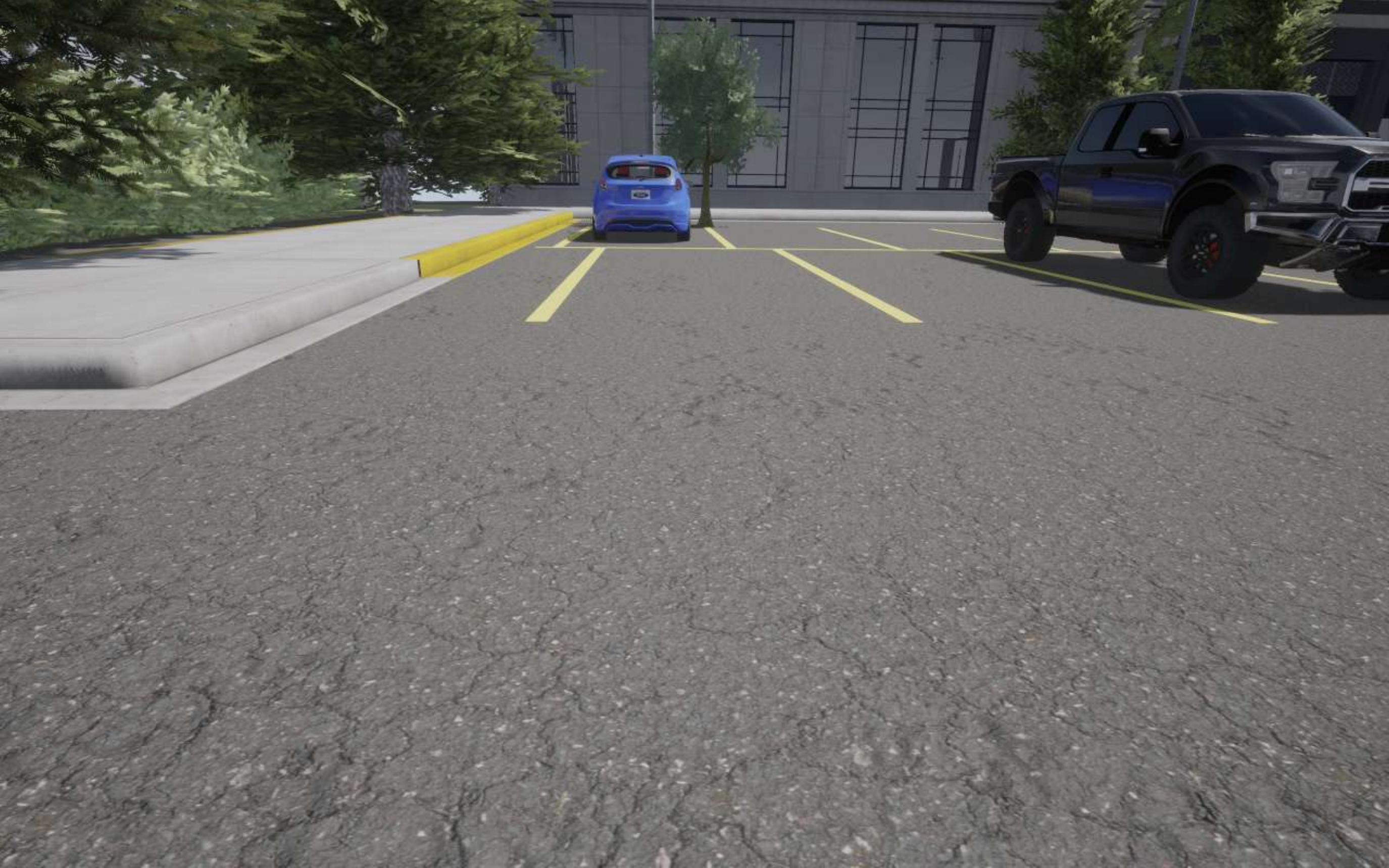}
        \caption{\small Simulated}
        \label{fig:sim_slot}
    \end{subfigure}
    \begin{subfigure}{.45\linewidth}
        \centering
        \includegraphics[height=.925in]{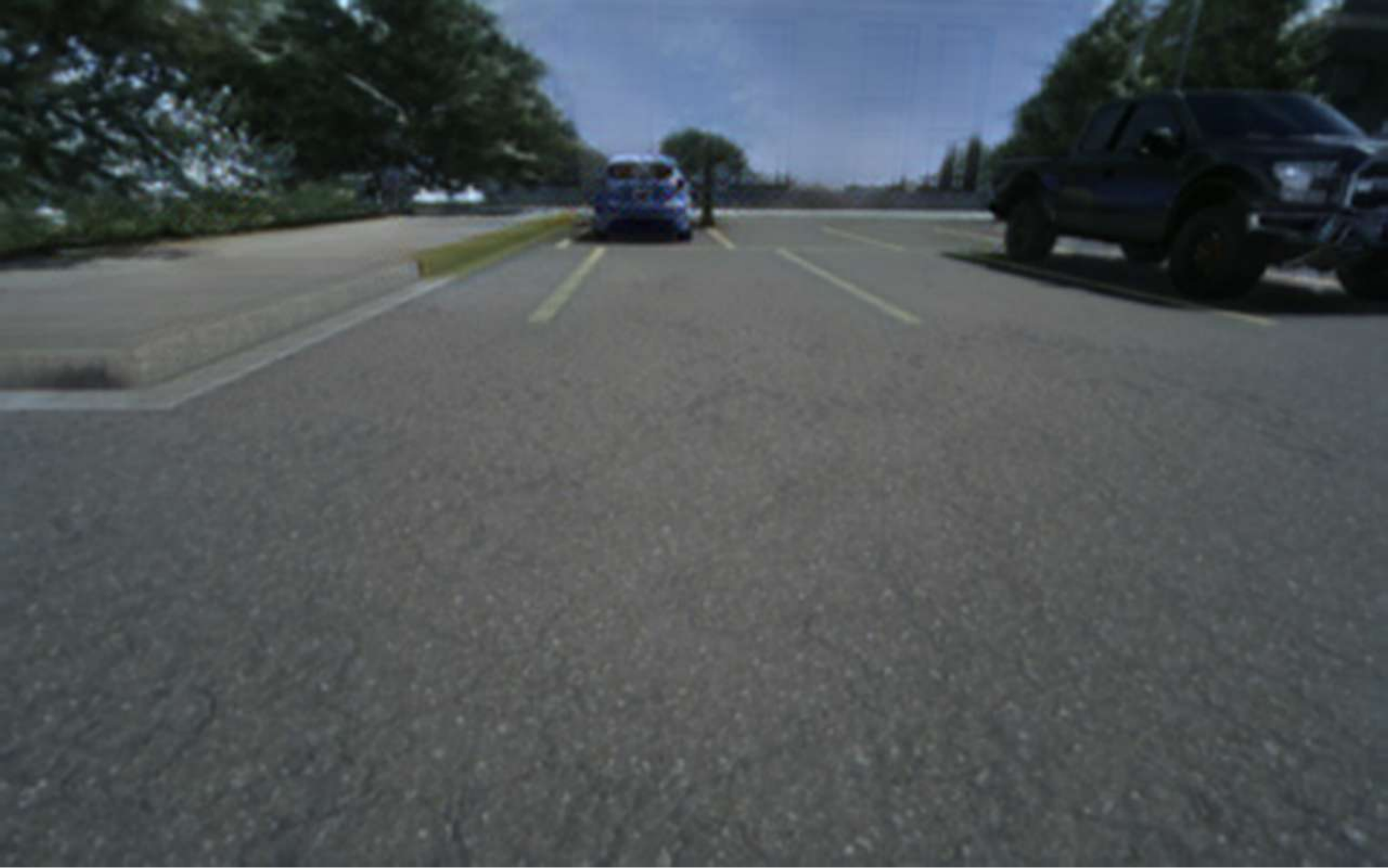}
        \caption{\small Sim2Real Translated}
        \label{fig:gan_slot}
    \end{subfigure}
\caption{Example images from the real and synthetic data used for the slot detection experiments.\label{fig:data_slot}}
\vspace{-.1in}
\end{figure}

\textbf{Real Data:} An internal parking dataset of bright daytime scenarios from two different parking lots (in Dearborn and Palo Alto) is used as the source of real data for this task. The Dearborn dataset has a total of $5907$ images, for brevity, we will refer to this dataset as \emph{Parking A}. The Palo Alto dataset has $602$ images. We will refer to this dataset as \emph{Parking B.} Fig.~\ref{fig:dearborn_slot} and Fig.~\ref{fig:paloalto_slot} show example images from the Parking A and Parking B datasets respectively to further motivate the large domain gap between them.
\newline
\textbf{Synthetic Data:} Simulated data for this task is generated using an Unreal Engine\footnote{\url{https://www.unrealengine.com/en-US/}}-based simulation pipeline for a diverse set of noise factors such as different times of the day, cloud density, shadow intensity/cast location, ground textures, parking line damage levels and parking density. The variety of shadow intensities and locations, along with parking line damage and car density are in stark contrast to the homogeneity of the parking A dataset. Fig.~\ref{fig:sim_slot} shows an example simulated image, visualizing the large domain gap between the simulated and real data from parking A. A sim2real VAE-GAN model (based on \cite{liu2017unsupervised,jaipuria2020role}) trained on unpaired simulated images and real images from the Parking A dataset is used to translate the generated simulated data to look photorealistic. Fig.~\ref{fig:gan_slot} shows the sim2real translated output for Fig.~\ref{fig:sim_slot}. Note the realistic ground textures and lighting effects in Fig.~\ref{fig:gan_slot} in contrast to Fig.~\ref{fig:sim_slot}.

\begin{figure}[h]
\centering
\includegraphics[width=.85\linewidth]{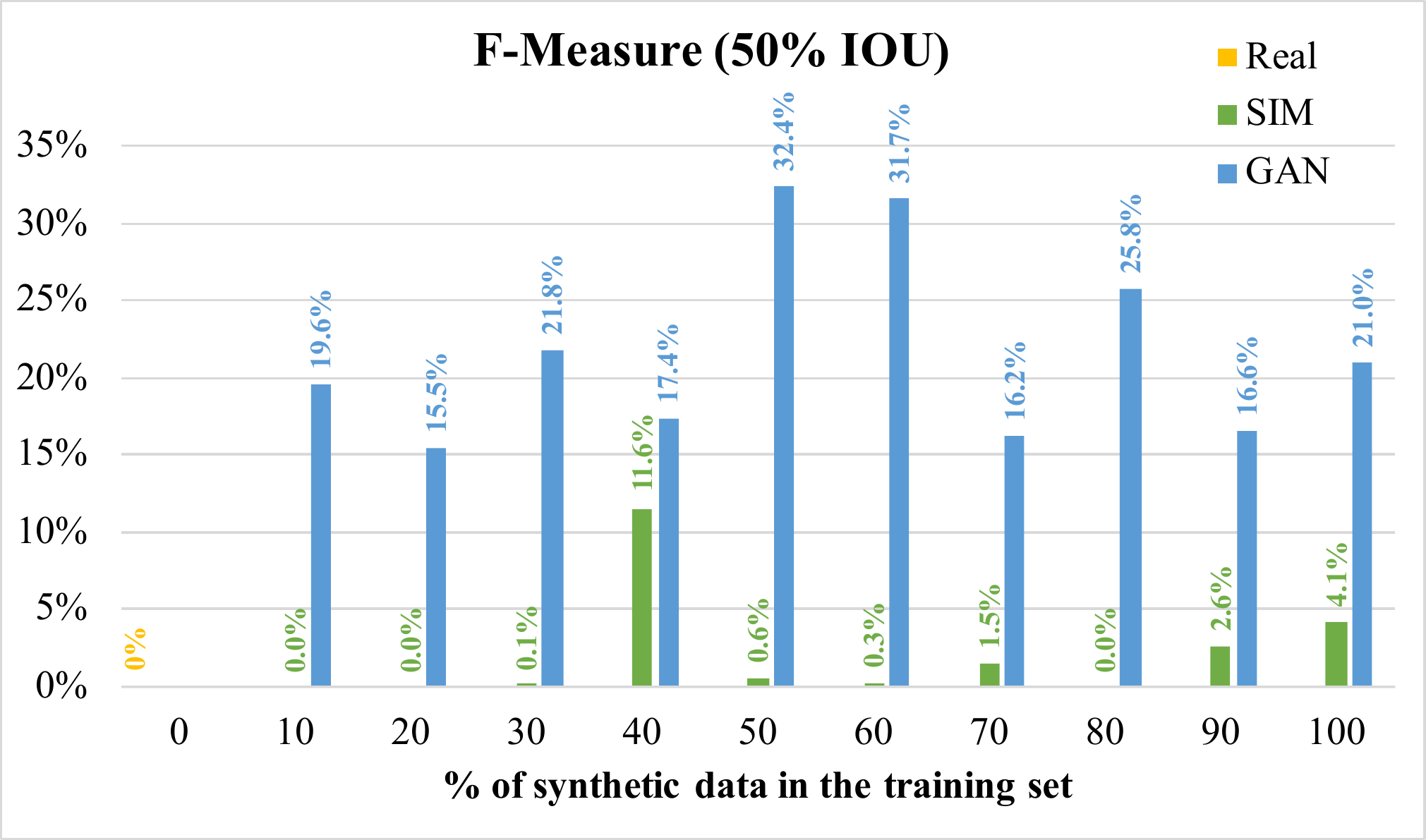}
\caption{Plot of F-measure for slot detection models trained on a mix of real (Parking A) and synthetic images (either from simulation or from sim2real GAN) and tested on real Parking B images. As you move from left to right, the ratio of synthetic data in the training set increases.\label{fig:F_slot}}
\vspace{-.2in}
\end{figure}

For the slot detection experiments in this paper, MobileNetV2 SSD \cite{sandler2018mobilenetv2,liu2016ssd}, pre-trained on COCO \cite{lin2014microsoft}, was trained and tested on $300\times300$ parking lot images to detect open parking slots, as shown in Fig.~\ref{fig:slot_detection}. The Parking A dataset was split into a train and test set with $3545$ images and $2362$ images respectively. Given the small size of the Parking B dataset (602 images), it was used for testing only. Intersection over Union (IoU) of detected slots with ground truth empty slots is used as the metric for quantitative evaluation. Post training, model checkpoint with the best F-measure for 50\% IoU on the Parking A test set is used for inference. The rest of this section describes the experiments performed to test our motivating hypothesis \texttt{H}.

\subsection{Results}
\label{sec:slot_results}
Fig.~\ref{fig:F_slot} shows the results of all slot detection models on the Parking B test set. Notice models trained on a mix of real and synthetic data (green and blue) significantly outperform the model trained on real data only (yellow). Moreover, across all ratios, models trained on a mix of real Parking A images and sim2real translated images (blue) do better than the models trained on a mix of real Parking A images and corresponding simulated images from Unreal Engine (green). Overall best performance (F-measure of 32.4\%) is achieved by the model trained on a mix of real and GAN data in a 50:50 ratio. Table~\ref{tab:experiments_slot} summarizes the results from the plots in Fig.~\ref{fig:F_slot}. For the synthetic data augmentation experiments, results are shown for the best model in terms of F-measure on cross-dataset testing. Additional insights into the number of true positives and false positives for cross-dataset testing with the models from Table~\ref{tab:experiments_slot} are provided in the Supplementary Material.

\begin{table}[h]
\footnotesize
  \centering
  \caption{Summary of results in Fig.~\ref{fig:F_slot}. Here, A and B denote the Parking A and Parking B datasets. S denotes simulated images and G denotes the sim2real translated equivalent of S. For synthetic data augmentation rows, results are shown for the best model in terms of F-measure on cross-dataset testing in green for A + S and in blue for A + G.}
  \vspace{-.1in}
    \begin{tabular}{|c|c|c|c|c|}
    \toprule
    \textbf{Train} & \textbf{Test} & \textbf{Precision} ($\uparrow$) & \textbf{Recall} ($\uparrow$) & \textbf{F-Measure} ($\uparrow$)\\
    \midrule
    A & A & 95.1\% & 87.9\% & 91.4\%\\
    $\text{A}+\text{S}\ (40\%)$ & A & 93.8\% & 87.7\% & 90.7\% \\
    $\text{A}+\text{G}\ (50\%)$ & A & 94.2\% & 86.5\% & 90.2\% \\
    \midrule
    A & B & 0\% & 0\% & 0\%\\
    $\text{A}+\text{S}\ (40\%)$ & B & \textcolor{green}{\textbf{71.8\%}} & \textcolor{green}{\textbf{6.3\%}} & \textcolor{green}{\textbf{11.6\%}} \\
    $\text{A}+\text{G}\ (50\%)$ & B & \textcolor{blue}{\textbf{67.0\%}} & \textcolor{blue}{\textbf{21.4\%}} & \textcolor{blue}{\textbf{32.4\%}} \\
    \bottomrule
    \end{tabular}%
  \label{tab:experiments_slot}%
  \vspace{-.15in}
\end{table}%
\subsection{Experiment Details}
\label{sec:exp_details_slot}
As shown in Table~\ref{tab:experiments_slot}, MobileNetV2 SSD trained on Parking A results in a F-Measure of 91.4\% on the Parking A test set (1\textsuperscript{st} row). However, the same model when tested on the Parking B dataset results in a F-measure of 0\% (4\textsuperscript{th} row). It is a well known fact that supervised learning-based methods do not generalize across different domains. In this particular case the generalization performance is much worse than one might expect because of two main reasons: (i) the small size (relative to large-scale image datasets such as ImageNet \cite{deng2009imagenet} and COCO \cite{lin2014microsoft}) and low diversity (all daytime images from the same parking lot) of the Parking A dataset; (ii) the large domain gap between the two datasets. Increasing dropout regularization did not help improve generalization performance either - F-Measure remained constant at 0\% for varying levels of dropout. The only improvement observed was in the number of false positives (more details are provided in Supplementary Material).

Thus, these results are consistent with the motivating hypothesis \texttt{H}. Additionally, as shown in the 2\textsuperscript{nd} and 3\textsuperscript{rd} rows of Table~\ref{tab:experiments_slot}, synthetic data augmentation did not adversely affect the results on the Parking A test set which further strengthens the case for the use of synthetic data and especially GAN-translated data to enrich real-world datasets for supervised learning tasks. 
\section{Case Study: Traffic Lane Detection}
\label{sec:lane}
\begin{figure}[h]
\centering
\includegraphics[width=.85\linewidth]{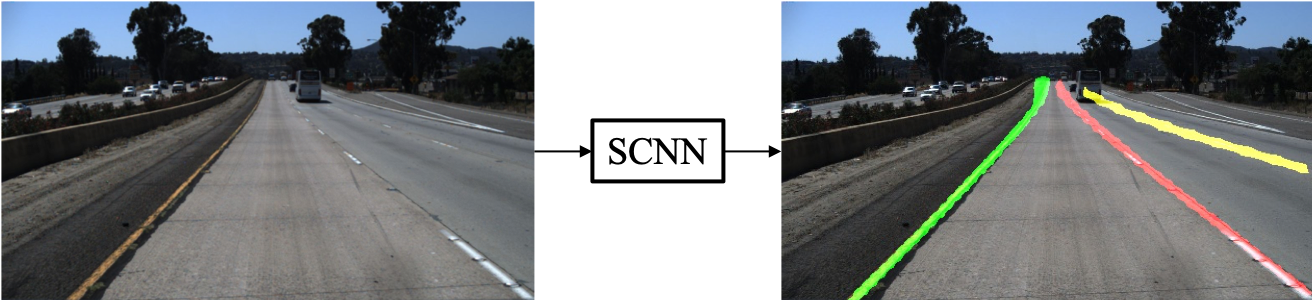}
\caption{Lane detection schematic.\label{fig:lane_detection}}
\vspace{-.15in}
\end{figure}

The objective of this task is to detect lane boundaries in images taken from a front vehicle camera (see Fig.~\ref{fig:lane_detection}).
Pan~\etal \cite{pan2018spatial} achieved state-of-the-art performance on this task with Spatial Convolutional Neural Networks (SCNNs). Their formulation is used as-is for all the lane detection experiments in this paper.

\subsection{Dataset Description}
\label{sec:data_lane}
\textbf{Real Data:} Following Pan ~\etal in \cite{pan2018spatial}, the \emph{CULane} and \emph{TuSimple}\footnote{\url{https://github.com/TuSimple/tusimple-benchmark/tree/master/doc/lane_detection}} datasets are used as real-world data sources. The CULane dataset has 88880 training images, 9675 validation images and 34680 test images - collected across diverse scenarios including urban, rural and highway environments. The TuSimple dataset has 3268, 358, and 2782 images for training, validation and testing respectively. Compared to CULane, TuSimple has highway scenes only.

\begin{figure}[h]
\centering
\includegraphics[width=1\linewidth]{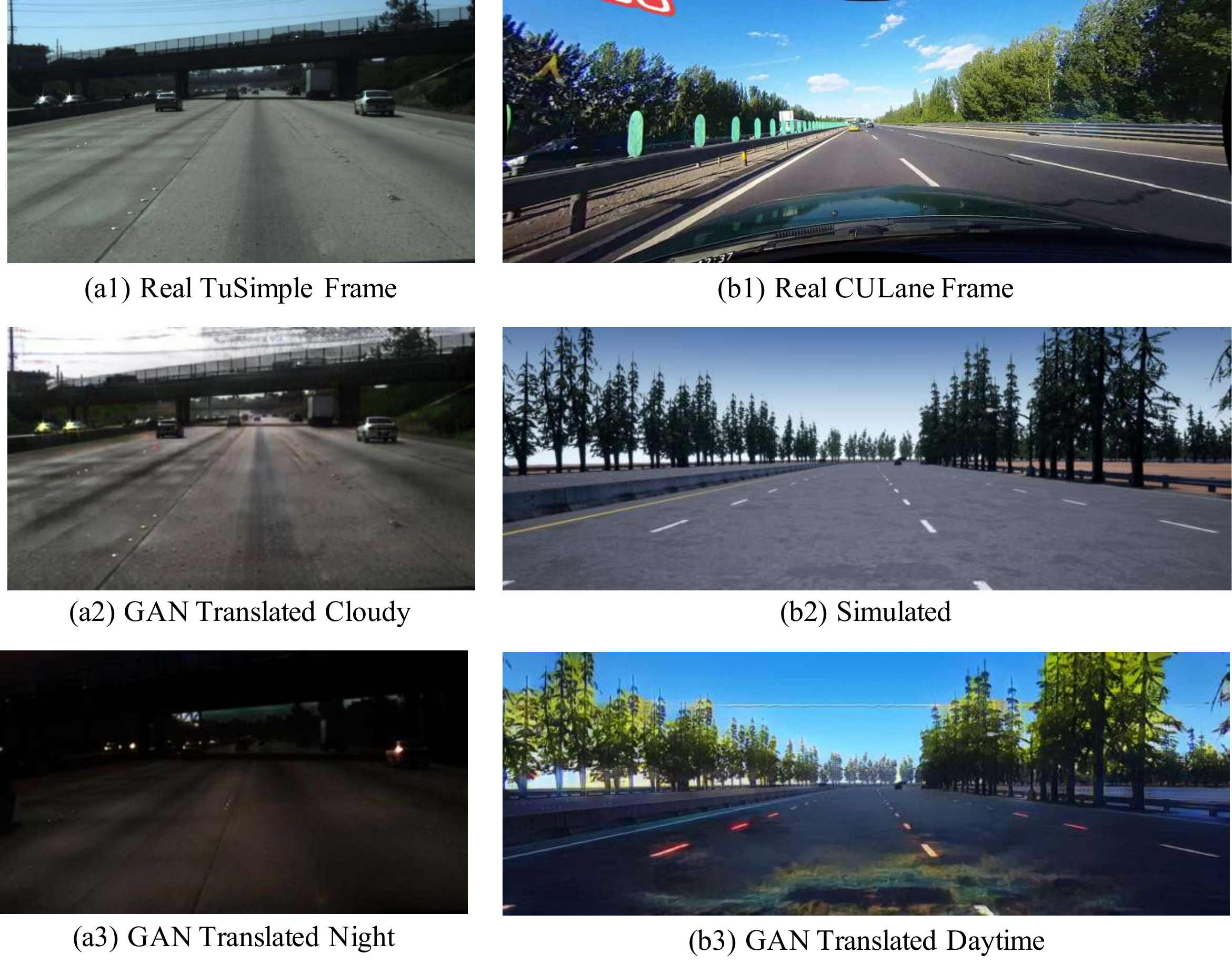}
\caption{Example real, simulated and GAN-translated images used for lane detection.
\label{fig:real_sim_gan_lane}}
\vspace{-.1in}
\end{figure}
\textbf{Synthetic Data:} For augmenting CULane, 88880 daytime highway images were generated using Unreal Engine by varying several noise factors such as the number of lanes, traffic density, sun intensity, location and brightness, road curvature, lane marker wear and tear etc. In testing the original implementation of SCNN, we found that the model performed poorest when lane lines were faint, in shadows or occluded by other vehicles. The change in sun intensity, its location and brightness helped create different shadow effects around the lane lines, giving the network more diverse data to train on. Varying traffic density and road curvature allowed for different occlusions of the lane line markings to produce more diverse data. Example synthetic images generated for this task are shown in Fig.~\ref{fig:real_sim_gan_lane}. Following the method outlined in Section~\ref{sec:method}, a sim2real VAE-GAN model (based on \cite{liu2017unsupervised,jaipuria2020role}) trained on unpaired simulated images and real images from CULane was used to translate the generated simulated data to look photorealistic. Fig.~\ref{fig:real_sim_gan_lane} shows the sim2real translated output for the given simulated image. Note the realistic ground textures and lighting effects in the GAN image in contrast to the simulated image.
\begin{figure}[h]
\centering
\includegraphics[width=.85\linewidth]{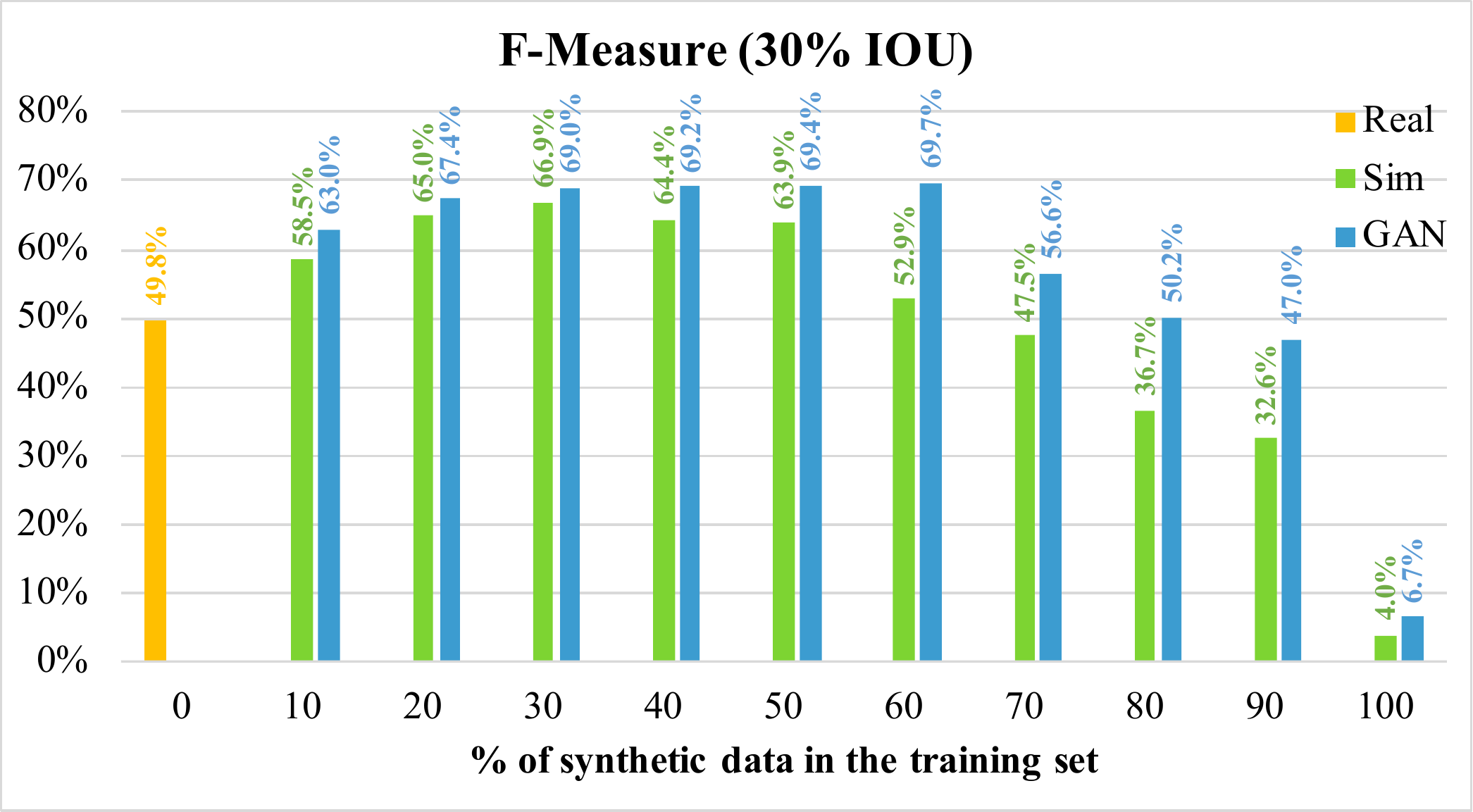}
\caption{Plot of F-measure for models trained on a mix of CULane and synthetic images (from simulation or from sim2real VAE-GAN) and tested on TuSimple images.\label{fig:fmeasure_cross_test_1}}
\vspace{-.35in}
\end{figure}

\subsection{Experiment Details}
For the lane detection experiments in this paper, two types of experiments were performed:\\
\textbf{Experiment I:} Following Section~\ref{sec:method}, SCNN \cite{pan2018spatial} is trained on a mix of CULane and synthetic images and tested on TuSimple. For results from SCNN trained on a mix of TuSimple and synthetic images and tested on CULane, please refer Supplementary Material. Models are trained on $800\times288$ images. For cross-testing, TuSimple images are padded (along width) to match the training resolution of $800\times288$ while simultaneously maintaining the original aspect ratio. IoU of detected lane lines with ground truth lane lines is used as the metric for quantitative evaluation.\\
\textbf{Experiment II:} In addition to the experiments described in Section~\ref{sec:method}, given that the TuSimple dataset has only daytime images while the CULane dataset has a diverse set of weather and lighting conditions (refer Section~\ref{sec:data_lane}), we performed an additional set of experiments for this task to further test the motivating hypothesis \texttt{H} particularly in scenarios where synthetic data augmentation addresses the specific bias of weather and lighting effects. All synthetic data was generated by applying day-to-night and clear-to-cloudy VAE-GAN models (based off of the architecture in Ref. \cite{liu2017unsupervised} and trained on BDD100K \cite{yu2018bdd100k}) to TuSimple images. Fig.~\ref{fig:real_sim_gan_lane} shows an example GAN night and cloudy image. SCNN was trained on $512\times288$ images for this set of experiments and tested on downsized and then padded (along height) versions of CULane images that match the training resolution of $512\times288$ while simultaneously maintaining the original aspect ratio.

\begin{table}[h]
\footnotesize
  \centering
  \caption{Summary of results in Fig.~\ref{fig:fmeasure_cross_test_1}. Here, A, A\textsubscript{N} and B denote the CULane, CULane Night only and TuSimple datasets. S denotes simulated images and G denotes the sim2real translated equivalent of S. G\textsubscript{N} and G\textsubscript{C} denote real TuSimple images translated to nighttime and cloudy respectively. For synthetic data augmentation rows, results are shown for the best model in terms of F-measure on cross-dataset testing in green for A + S and in blue for A + G.}
  \vspace{-.1in}
    \begin{tabular}{|c|c|c|c|c|}
    \toprule
    \textbf{Train} & \textbf{Test} & \textbf{Precision} ($\uparrow$)& \textbf{Recall} ($\uparrow$)& \textbf{F-Measure} ($\uparrow$)\\
    \midrule
    A & A & 53.6\% & 70.6\% & 60.9\%\\
    $\text{A}+\text{S}\ (30\%)$ & A & 53.5\% & 70.8\% & 60.9\% \\
    $\text{A}+\text{G}\ (60\%)$ & A & 51.6\% & 68.0\% & 58.7\% \\
    \midrule
    A & B & 47.8\% & 51.9\% & 49.8\%\\
    $\text{A}+\text{S}\ (30\%)$ & B & \textcolor{green}{\textbf{63.6\%}} & \textcolor{green}{\textbf{70.6\%}} & \textcolor{green}{\textbf{66.9\%}} \\
    $\text{A}+\text{G}\ (60\%)$ & B & \textcolor{blue}{\textbf{67.8\%}} & \textcolor{blue}{\textbf{71.6\%}} & \textcolor{blue}{\textbf{69.7\%}}\\
    \midrule
    \midrule
    B & B & 80.2\% & 91.7\% & 85.6\%\\
    $\text{B}+\text{G}\textsubscript{N}\ (10\%)$ & B & 80.3\% & 91.9\% & 85.7\%\\
    $\text{B}+\text{G}\textsubscript{C}\ (80\%)$ & B & 79.4\% & 90.6\% & 84.7\% \\
    \midrule
    B & A & 2.8\% & 3.7\% & 3.2\% \\
    $\text{B}+\text{G}\textsubscript{N}\ (10\%)$ & A & 5.9\% & 7.8\% & 6.7\% \\
    $\text{B}+\text{G}\textsubscript{C}\ (80\%)$ & A & \textbf{6.6\%} & \textbf{8.7\%} & \textbf{7.5\%} \\
    \midrule
    \text{B} & A\textsubscript{N} & 0.2\% & 0.3\% & 0.2\% \\
    $\text{B}+\text{G}\textsubscript{N}\ (10\%)$ & A\textsubscript{N} &  2.3\% & 3.1\% & 2.6\% \\
    \bottomrule
    \end{tabular}
  \label{tab:experiments_lane_a2b}
\vspace{-.2in}
\end{table}

\subsection{Results}
\textbf{Experiment I:} Consistent with the cross-testing results in Section~\ref{sec:slot}, as shown in Table~\ref{tab:experiments_lane_a2b}, SCNN trained on CULane results in a F-Measure of 60.9\% on the CULane test set (1\textsuperscript{st} row) versus 49.8\% on the TuSimple test set (4\textsuperscript{th} row). This drop in accuracy can again be attributed to the large domain gap between the two datasets (see Fig.~\ref{fig:real_sim_gan_lane}). Fig.~\ref{fig:fmeasure_cross_test_1} shows that models trained with a mix of real and sim2real translated data (blue) consistently outperform models trained with a mix of real and simulated data (green) in cross-testing. Moreover, as the ratio of synthetic data in the training set increases, the gap between models trained on GAN data and simulated data grows wider. Both these observations together verify the closeness of the GAN data to the real data as compared to just simulated data. More interestingly, for certain ratios of synthetic data, the models trained on a mix of real and synthetic data significantly outperform models trained with 100\% real data. Table~\ref{tab:experiments_lane_a2b} (top) summarizes the results from the best models in terms of F-Measure - 69.7\% for model trained on a 40:60 mix of real and GAN data and 66.9\% for model trained on a 70:30 mix of real and sim data versus just 49.8\% for model trained on 100\% real data (note the size of the training dataset was held constant across all experiments). These results confirm that synthetic data augmentation can help deflate dataset bias and thus improve cross-dataset generalization performance. Again, similar to the observations in Section~\ref{sec:slot_results}, the drop in accuracy on the original test set is minimal.
\\
\textbf{Experiment II:} Consistent with previous results, SCNN trained on TuSimple gives an F-measure of 85.6\% on the TuSimple test set versus only 3.2\% on the CULane test set (7\textsuperscript{th} row vs. 10\textsuperscript{th} row in Table~\ref{tab:experiments_lane_a2b}). The drop in accuracy is more prominent in this case as TuSimple is a much simpler dataset as compared to CULane both in terms of quantity and diversity. Table~\ref{tab:experiments_lane_a2b} shows that adding nighttime and cloudy data helps improve cross-dataset generalization performance, with models trained on a mix of real and GAN-generated cloudy data faring the best among all (12\textsuperscript{th} row in green). Since CULane had the nighttime images labeled in their test set, we compared the performance of models trained on TuSimple only with models trained on a mix of TuSimple and GAN nighttime images and again, consistent with our motivating hypothesis \texttt{H}, the latter models do better (last row).
\section{Case Study: Monocular Depth Estimation}
\label{sec:depth}
\vspace{-.1in}
\begin{figure}[h]
\centering
    \begin{subfigure}{.4\textwidth}
        \centering
        \includegraphics[width=1\linewidth]{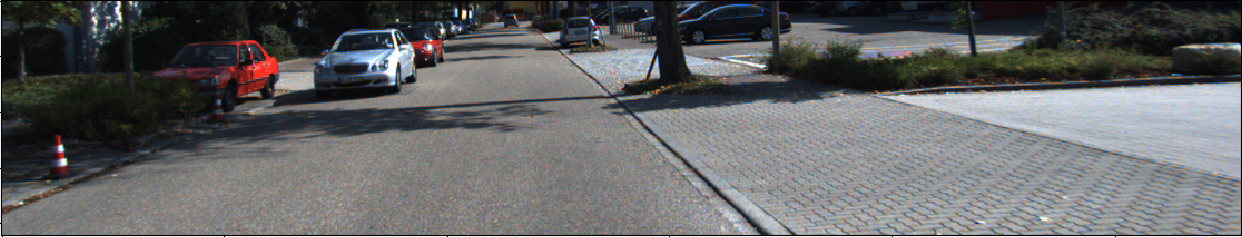}
    \end{subfigure}
    \begin{subfigure}{.4\textwidth}
        \centering
        \includegraphics[width=1\linewidth]{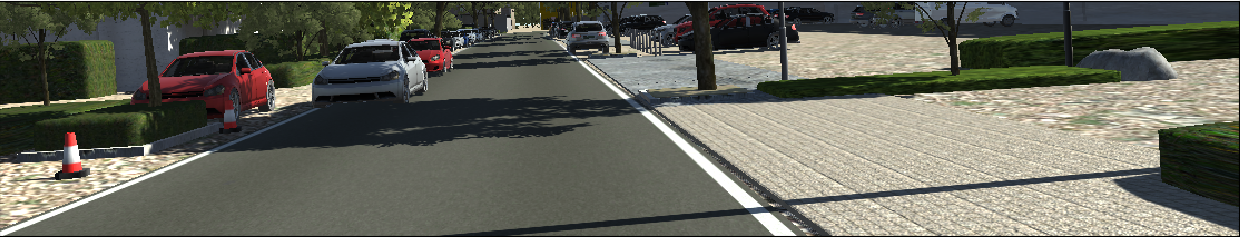}
    \end{subfigure}
    \begin{subfigure}{.4\textwidth}
        \centering
        \includegraphics[width=1\linewidth]{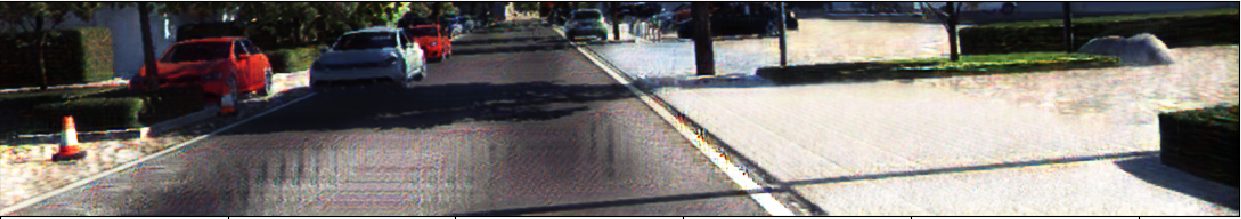}
    \end{subfigure}
    \begin{subfigure}{.4\textwidth}
        \centering
        \includegraphics[width=1\linewidth]{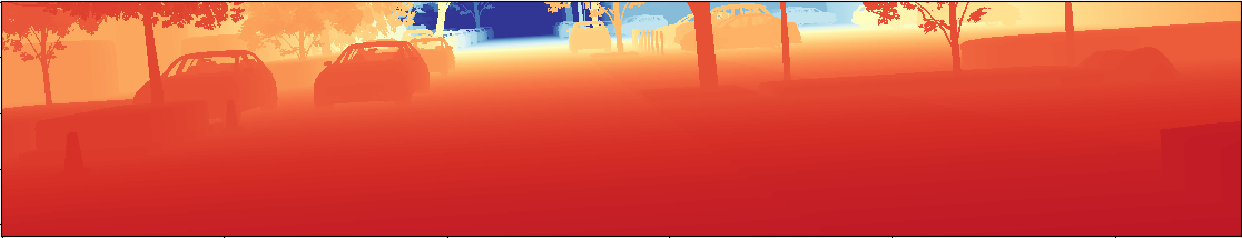}
    \end{subfigure}
    \begin{subfigure}{.4\textwidth}
        \centering
        \includegraphics[width=1\linewidth]{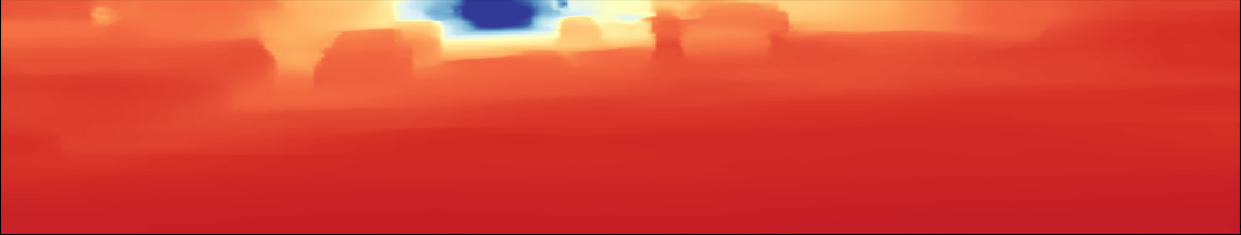}
    \end{subfigure}
    \begin{subfigure}{.4\textwidth}
        \centering
        \includegraphics[width=1\linewidth]{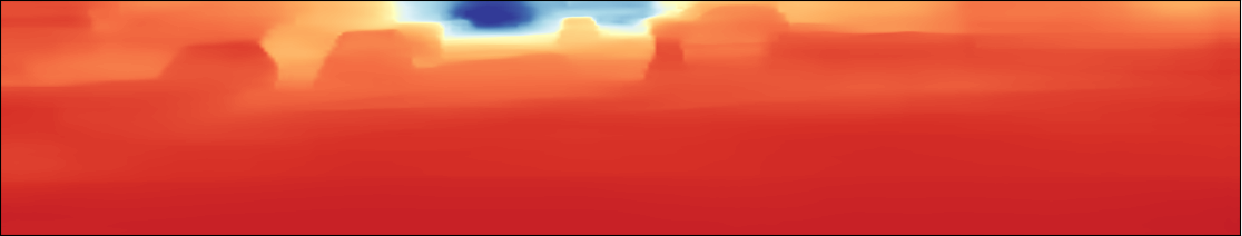}
    \end{subfigure}
        \begin{subfigure}{.4\textwidth}
        \centering
        \includegraphics[width=1\linewidth]{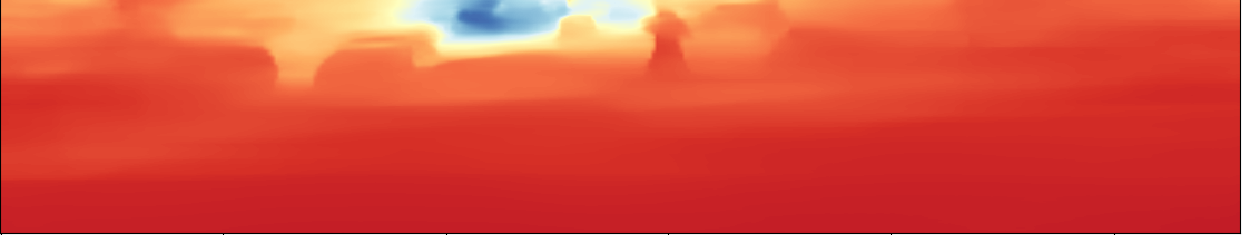}
    \end{subfigure}
\caption{From top to bottom: KITTI RGB, vKITTI RGB, sim2real, ground truth depth, estimated depth A+S ($60 \%$), estimated depth A+G ($60 \%$) and estimated depth A+G ($20 \%$). Networks were trained with unpaired data. Paired images are used for illustrative purposes only. 
\label{fig:single_image_depth}}
\vspace{-.1in}
\end{figure}

In this case study, experiments are conducted for the task of estimating the depth in a scene from a single RGB image ~\cite{eigen2014depth, garg2016unsupervised, godard2017unsupervised, zhou2017unsupervised}. We employ an encoder-decoder architecture with skip connections and train the network in a supervised fashion with MSE and edge-aware losses \cite{godard2017unsupervised} between the ground truth and estimated depth maps.

\subsection{Dataset Description}
We use KITTI \cite{geiger2012we} and virtual KITTI (vKITTI) \cite{gaidon2016virtual} as our real and simulated datasets. The vKITTI dataset  is a scene-by-scene recreation of the KITTI tracking dataset, also using the Unreal gaming engine. However, we don't use any paired data for our experiments. We also do not use data from the same sequences as the real data for our simulated data.
\newline
\textbf{Real Data:} 
We use the KITTI odometry sequence 00, with a total of 4,540 images as our real training set - \textbf{A}. The KITTI Odometry sequences 02 and 05, with a cumulative 500 images, are used as the real test set - \textbf{B}. Ground truth depth is generated by using the OpenCV implementation of the stereo algorithm SGBM with WLS filtering on the left and right images. Note that since we did not make use of paired images between the simulated and real datasets, we could not use simulated depth as ground truth. Moreover, while the simulated recreation in vKITTI approaches that of real KITTI, the simulacrum is not exact, and this would have resulted in systematic biases in the learning of depth. This can be seen in rows 1 and 2 (KITTI and vKITTI) of Figure \ref{fig:single_image_depth}, where the virtual clone of the tree trunk on the right sidewalk is subtly different and slightly shifted.
\newline
\textbf{Synthetic Data:} 
We use data from vKITTI scenes 1, 2, 6, 18 and 20, under the Clone, Morning, 15L and 15R subsets, resulting in a total of 2,126 images per subset, and an overall total of 8,504 images. 
These vKITTI scenes are clones of the KITTI Tracking dataset (Clone), with variation in camera angles (15L/R) and time of the day (Morning).
Note that the KITTI Tracking sequences (duplicated in vKITTI) are captured in a different environment compared to the KITTI Odometry dataset, which form part of our Real set. This variation in sequence geographical location, time of the day and camera pan angles represent the noise factors for this task.
A set of randomly picked 4,540 images from this total is used as the source of simulated data for training - \textbf{S}. We use cycleGAN \cite{zhu2017unpaired}, trained with unpaired images from KITTI and vKITTI to convert the 4,540 sampled images from vKITTI to make them more realistic. This forms our sim2real translated dataset - \textbf{G}.\\
\vspace{-.2in}
\subsection{Experiment Details}
As with the other tasks, we train the task network with different percentages of simulated (\textbf{A + S}) and sim2real (\textbf{A + G}) data, starting from $0\%$ to $100\%$ and test on KITTI sequences that were not seen during training (\textbf{B}). We use the Root Mean Squared Error (RMSE) metric to determine the performance of the network trained on a particular sim/real or sim2real/real mix, after limiting maximum depth to 100m. 
We provide detailed RMSE results in Figure \ref{fig:monodepth_metrics_rmse}. We also tested this task based on accuracy of depth estimation, measured as the ratio of correctly estimated depth pixels to the total number of depth pixels. These results are summarized, along with RMSE in Table \ref{tab:experiments_depth} and more detailed results for accuracy are provided in the Supplementary Material. RMSE and accuracy are common metrics used in prior work on single image depth \cite{eigen2014depth}. A lower value of RMSE indicates better performance while the same is true for a higher value for accuracy.

\subsection{Results}
\begin{center}
\begin{figure}[h]
\centering
    \includegraphics[width=.85\linewidth]{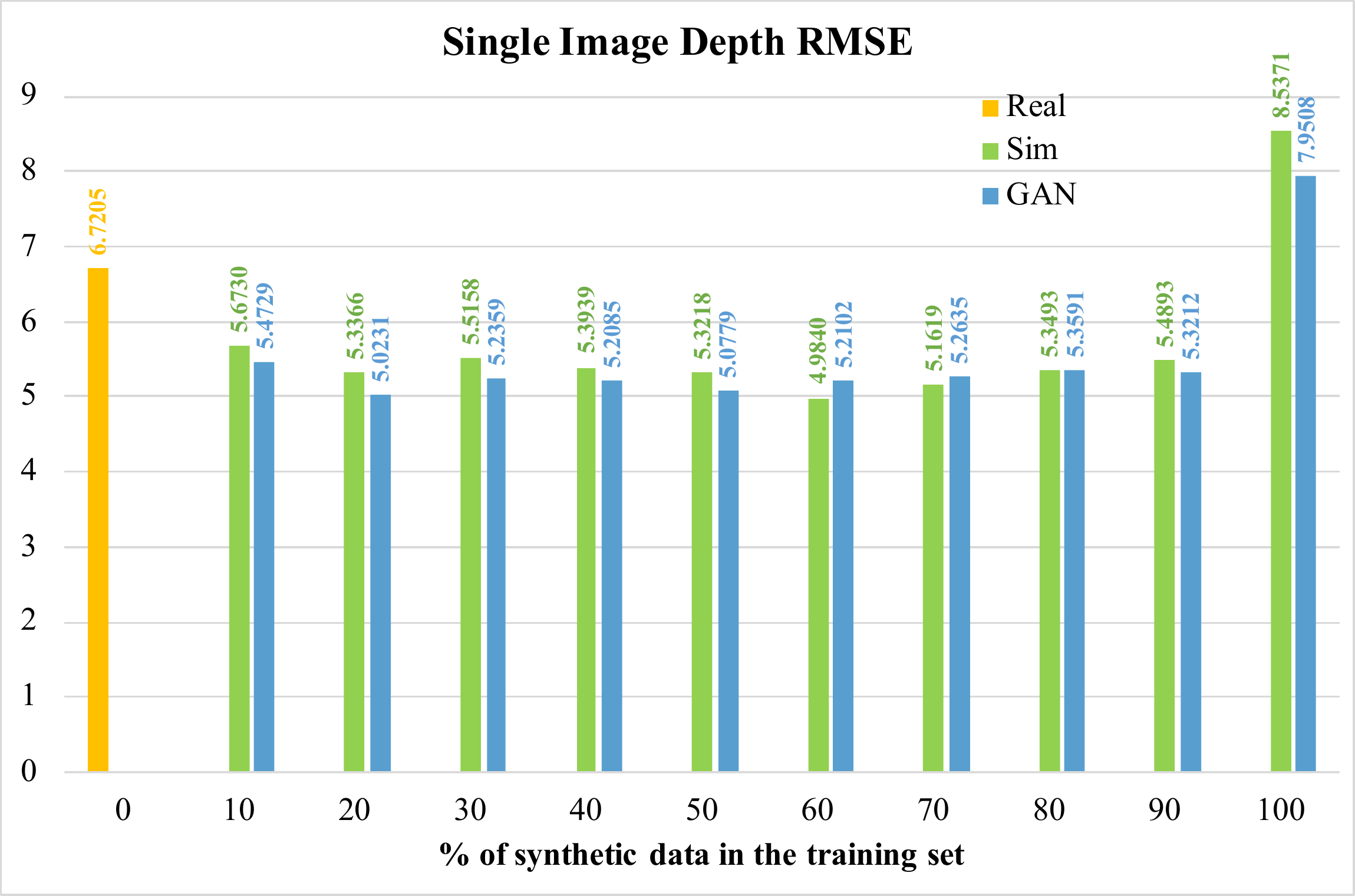}
   \caption{RMSE results for the single image depth task (lower is better).}
\vspace{-.35in}
\label{fig:monodepth_metrics_rmse}
\end{figure}
\end{center}

Figure \ref{fig:monodepth_metrics_rmse} shows RMSE for the different mixes of real (yellow), real + simulated (A+S, Sim, green) and real + sim2real (A+G, GAN, blue) training data. Some important highlights of the same are shown in Table \ref{tab:experiments_depth}.
\vspace{-.1in}

From the RMSE numbers, one result is clear: having either simulated or sim2real data in the training mix is better than using only real data, for the same amount of total training data. This is shown by the yellow bar (only real data) being higher than the other mixes. Equally, having simulated (sim/sim2real) data alone (the last pair of bars in the RMSE figure) gives the worst results. The trends indicate that mixing sim2real (after converting the simulated data with the sim2real GAN pipeline) with real is better than mixing sim with real, when the percentage of sim/sim2real data is lower or equal to the percentage of real data ($10-50\%$ sim/sim2real), in the left half of the bar graphs. In other words, A + G seems to give a slight performance gain over A + S in the $10-50\%$ range. From Table \ref{tab:experiments_depth}, we see that the absolute best performer in terms of RMSE is $20\%$ A + G and $60\%$ A + S. Qualitative results are shown for a single image in Figure \ref{fig:single_image_depth}. Visually, the A + G ($60 \%$) network (trained with a 40/60 mix of real and sim2real data) seems to perform the best on this image, followed by A + S ($60 \%$). The top performer in terms of RMSE, A + G ($20 \%$) looks visually slightly worse.

\begin{table}[h]
\footnotesize
  \centering
  \caption{Summary of results for the single image depth task. Best results for A + S are in green, and best results for A + G are in blue.}
    \vspace{-.1in}
    \begin{tabular}{|c|c|c|c|}
    \toprule
    \textbf{Train} & \textbf{Test} & $\mathbf{RMSE}$ ($\downarrow$) & $\mathbf{ Accuracy}$($\uparrow$) \\
    \midrule
    \text{A} & \text{B} & 6.7205 & 0.9559 \\
    \text{A} + \text{S}\ (20\%) & B & 5.3366 & 0.9705  \\
    \text{A} + \text{G}\ (20\%) & B & \textcolor{blue}{\textbf{5.0231}} & 0.9712  \\
    \text{A} + \text{S}\ (50\%) & B & 5.3218 & 0.9702  \\
    \text{A} + \text{G}\ (50\%) & B & 5.0779 & \textcolor{blue}{\textbf{0.9721}}  \\
    \text{A} + \text{S}\ (60\%) & B & \textcolor{green}{\textbf{4.9840}} & \textcolor{green}{\textbf{0.9723}}  \\
    \text{A} + \text{G}\ (60\%) & B & 5.2102 & 0.9682  \\
    \bottomrule
    \end{tabular}
  \label{tab:experiments_depth}
  \vspace{-.2in}
\end{table}

Another important result to be highlighted is the fact that the network trained on just simulation data gains about $7\%$ in terms of RMSE with the sim2real transformation, when tested on real data when using the accuracy numbers. This shows that sim2real from simulation to the source dataset,
without any labelling from the source set, already gives a baseline boost. This, when mixed with real labelled data from the source set allows single image depth performance on the target set to rise further, and the perfect mix of real and simulated data lies in the 80/20 to 40/60 range, with sim2real showing minor improvements over just using simulated data in the mix. 

We also conducted single image depth experiments using the NuScenes dataset \cite{caesar2019nuscenes} for real data and the CARLA simulation environment \cite{dosovitskiy2017carla} for simulation data. These experiments indicated that the estimation of a depth map from a single image is highly dependent on the focal length and other intrinsic camera parameters. We were able to get good results on the NuScenes dataset by using data from CARLA, when the simulated camera on CARLA had been matched with the intrinsics the NuScenes camera. However, any mix of KITTI with NuScenes/CARLA  during training completely confounded the algorithm and we do not include these experiments in this paper. We consider camera intrinsics an important consideration when generating simulation and sim2real data and one has to match these with the target dataset. The mixing of data across datasets captured with different focal length cameras requires more sophisticated techniques that are beyond the scope of this paper.
\section{Discussion}
As motivated in Section~\ref{sec:intro}, dataset bias is a known problem in computer vision. However, most of the well studied sources of bias are task-agnostic. In this work, we focus on bias in the context of the noise distribution pertaining to task-specific environmental factors, referred to as \emph{noise factor distribution bias}, and show that targeted synthetic data augmentation can help deflate this bias. For empirical verification, we use three different computer vision tasks of immense practical use  - parking slot detection, lane detection and monocular depth estimation. Synthetic data for these tasks is generated via a simple two step process: (i) simulate images for a diverse set of task-specific noise factors and obtain corresponding ground truth; (ii) perform sim2real translation using GANs to make simulated images look like they are from the real training domain. The rest of this section summarizes the key insights obtained.

Across all three tasks, having synthetic data in the training mix provides a significant boost in cross-dataset generalization performance as compared to training on real data only, for the same size of the training set. Moreover, performance on the source domain test set was not adversely impacted which makes the case for synthetic data augmentation to enrich training datasets for these tasks stronger.

For both the slot detection and lane detection tasks, the best models in terms of F-Measure were those trained on a mix of real and sim2real translated data. For slot detection, the best model with 50\% sim2real data in the training mix provided about 30\% absolute improvement over the model trained on 100\% real data. For lane detection, the best model with 60\% sim2real data in the training mix performed about 40\% better than the one trained on 100\% real data. Another consistent observation across the two tasks is that models with a higher ratio of synthetic data ($>$ 50\%) in the training mix do much better when the source of the synthetic data is sim2real data as opposed to simulated data.

In contrast, for the depth estimation task, the best model in terms of both RMSE and accuracy was the one with 60\% simulated data (and not sim2real data) in the training mix that achieved a 25\% improvement in RMSE over the model trained with 100\% real data. We think this is because of two main reasons. First, depth estimation from a sensor (RGB camera) that is missing the 3\textsuperscript{rd} dimension is an inherently hard task with every pixel contributing to the error metric. If we were solving some other problem in which 3D estimation can be parameterized - \eg 3D bounding box detection from 2D images - instead of requiring prediction on a pixel level, we would expect to see a bigger gain with sim and sim2real data added in the training mix. Secondly, slot detection and lane detection are mostly dependent on higher-level features (such as edges) and appearance (such as exposure and lighting conditions). Sim2real is good at doing exactly this - matching higher-level features between the generated and real images and thus these two tasks significantly benefit from sim2real. Depth estimation, however, is dependent more on low-level features. Artifacts introduced by the GAN make it difficult to bridge the low-level feature discrepancies between the sim2real images and corresponding ground truth annotation obtained from simulation. Thus, as expected, for this task, as you go higher in terms of the ratio of synthetic data in the training mix ($>$ 50\%), models trained on a mix of real and simulated data do better than those trained on a mix of real and sim2real data. However, the model trained on 100\% sim2real data outperforms the one trained on 100\% simulated data for this task as well.

Another interesting finding is that across all three tasks, the best models in terms of the chosen metrics were always those with 50\%-60\% synthetic data in the training mix. Although this makes intuitive sense, it requires more in-depth investigation which will be part of future work.

{\footnotesize
\bibliographystyle{ieee_fullname}
\bibliography{ms}
}
\clearpage

\setcounter{section}{0}
\renewcommand{\thesection}{\Alph{section}}
\onecolumn
\section{Supplementary Material}
\begin{figure}[h!]
\centering
    \begin{subfigure}{.4\textwidth}
        \centering
        \includegraphics[width=1\linewidth]{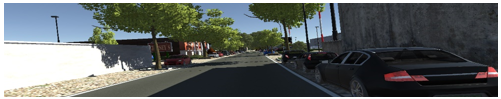}
    \end{subfigure}
    \begin{subfigure}{.4\textwidth}
        \centering
        \includegraphics[width=1\linewidth]{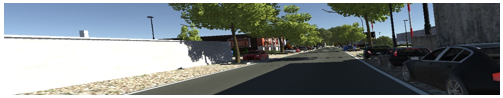}
    \end{subfigure}
    \begin{subfigure}{.4\textwidth}
        \centering
        \includegraphics[width=1\linewidth]{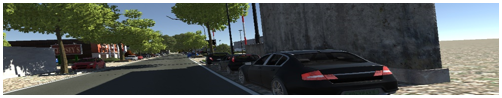}
    \end{subfigure}
    \begin{subfigure}{.4\textwidth}
        \centering
        \includegraphics[width=1\linewidth]{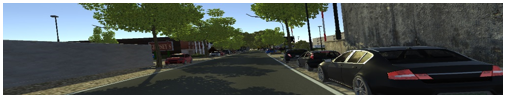}
    \end{subfigure}
\caption{From top to bottom: Images from vKITTI Clone, 15L, 15R and Morning subsets used as simulation data for the single image depth task. 
\label{fig:single_image_depth_noise_factors}}
\vspace{-.15in}
\end{figure}

Section~\ref{sec:noise_factor} of this supplementary material gives a deeper insight into the noise factor distribution of simulated data generated using Unreal Engine\footnote{\url{https://www.unrealengine.com/en-US/}} for the targeted synthetic data augmentation case studies of parking slot detection and traffic lane detection. Since simulated data used for the third case study of monocular depth estimation was sampled from the publicly available virtual KITTI \cite{gaidon2016virtual} dataset with source details provided in the main paper, we do not include any additional statistics here. Example simulated images used for the depth estimation task are shown in Fig.~\ref{fig:single_image_depth_noise_factors} in Section~\ref{sec:supp_quant}.

\begin{figure*}[h!]
\centering
\includegraphics[width=1\linewidth]{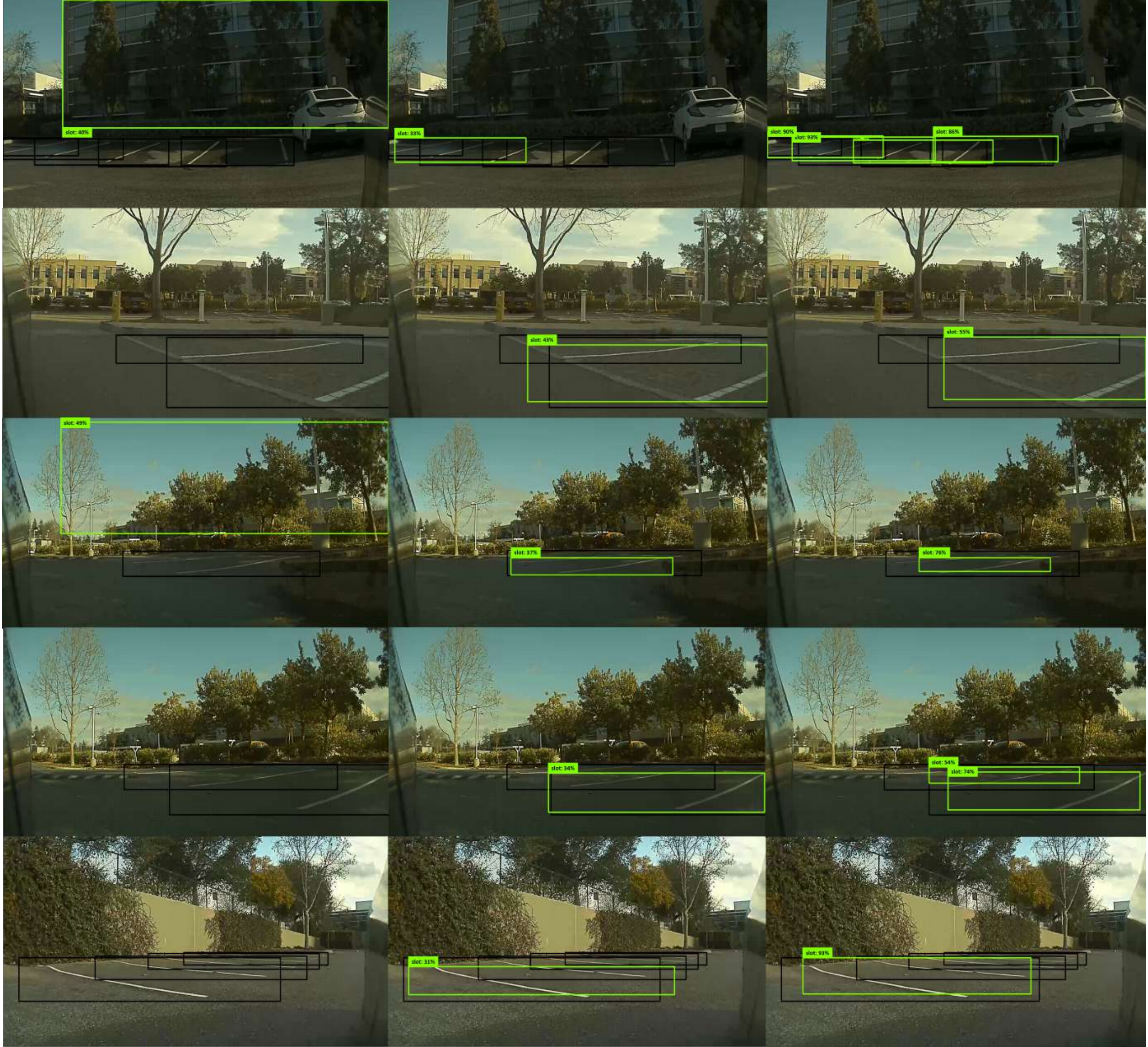}
\caption{Qualitative comparison of slot detection results on the held-out Parking B test set. Each row shows results on one example test image. Here, black boxes denote ground truth and green boxes are model predictions. The three columns from left to right show results from models trained on real Parking A dataset only, models trained on a mix of real and simulated data and models trained on a mix of real and sim2real data translated to look like Parking A real data. Note the number of true positives (TPs) and confidence scores increases from left to right. The first row shows how the number of TPs increases and the last row shows how confidence score goes from undetected to detected with a confidence improvement from 31\% to 93\% confidence for the same image. Confidence scores are best viewed by zooming into the relevant figure.\label{fig:slot_cross_qualitative}}
\end{figure*}

\begin{figure*}[h!]
\centering
\includegraphics[width=\textwidth,height=\textheight,keepaspectratio]{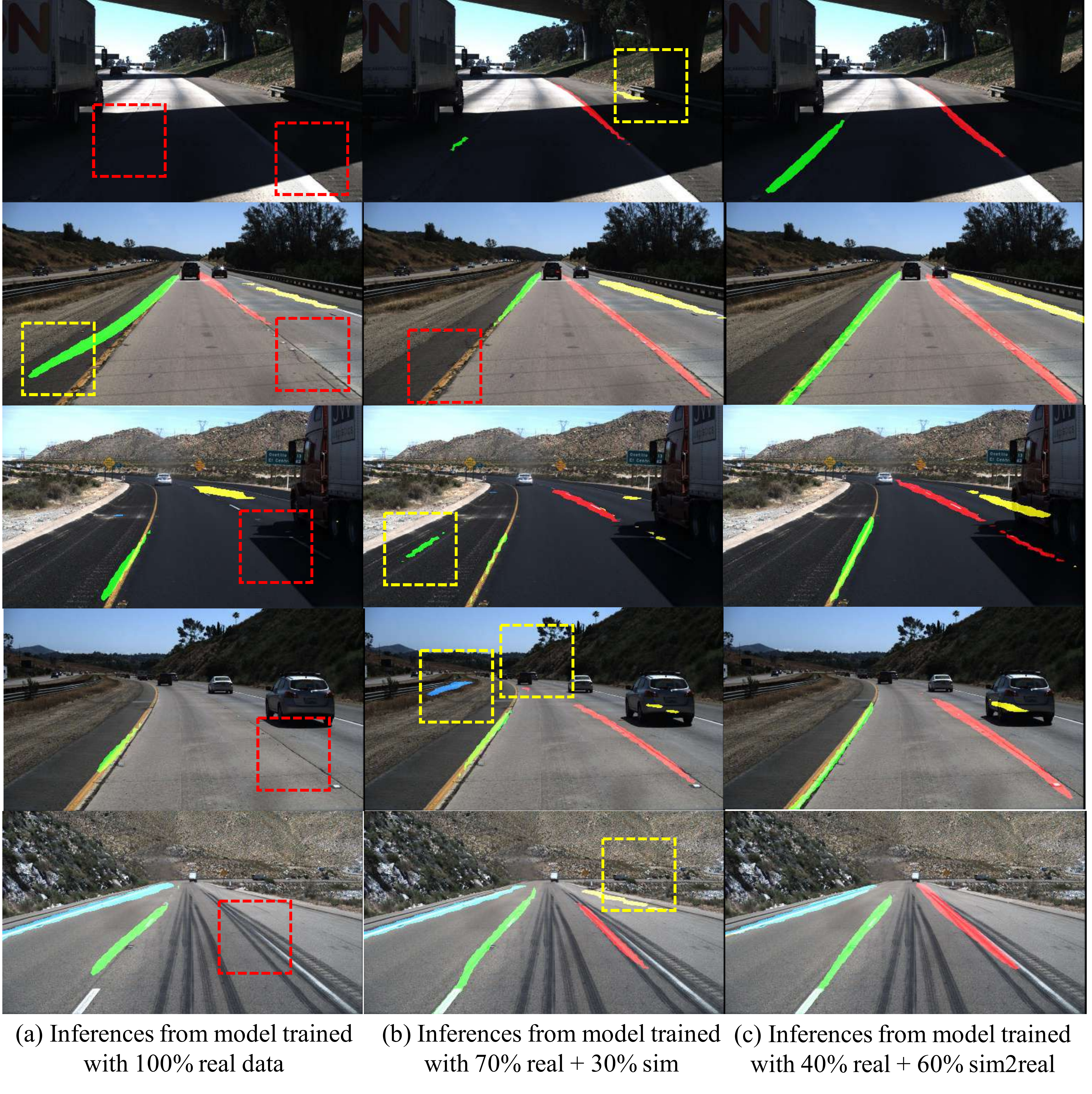}
\caption{Qualitative results from the cross-dataset generalization experiments (train on CULane, test on TuSimple) in the paper for the task of lane detection. Columns (a), (b) and (c) show the results from models trained on 100\% real data, 70\% real + 30\% sim data and 40\% real + 60\% sim2real data respectively. The red dashed squares highlight false negatives and the yellow dashed squares highlight false positives.\label{fig:lane_cross_qualitative}}
\end{figure*}

\begin{figure*}[h!]
\centering
\includegraphics[width=\textwidth,height=\textheight,keepaspectratio]{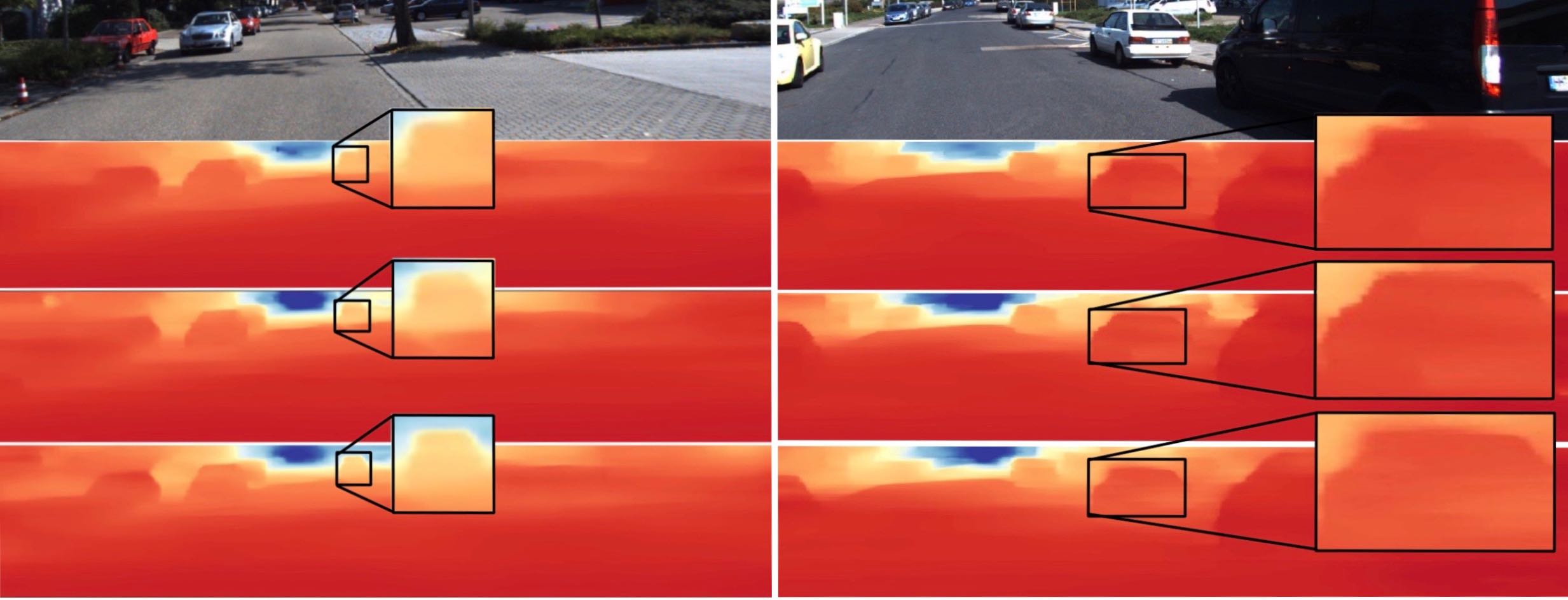}
\caption{Qualitative comparison of results from single-image depth estimation models trained on the KITTI Odometry Sequence 00 + Virtual KITTI datasets and tested on the KITTI Tracking dataset. Row 1 shows two different test images from KITTI tracking dataset, Row 2 shows the corresponding depth estimation results from a model trained on 100\% real data, Row 3 shows the depth estimation results from a model trained on 40\% real + 60\% sim and Row 4 shows the depth estimation results from a model trained on 40\% real + 60\% sim2real data.The addition of sim and sim2real data in the training mix (rows 3 and 4) improves the crispness of the depth estimation along vehicle boundaries as shown by the expanded insets.
\label{fig:rgb2depth_cross_qualitative}}
\end{figure*}

Qualitative results of targeted synthetic data augmentation are also included in this supplementary material in Sections~\ref{sec:qual_slot},~\ref{sec:qual_lane} and~\ref{sec:qual_depth} for the tasks of slot detection, lane detection and depth estimation respectively. Fig.~\ref{fig:slot_cross_qualitative} shows qualitative results of cross-dataset generalization experiments from the paper for the task of parking slot detection. Note the significant improvement in the number of true positives and their confidence scores as we move from left to right with the leftmost column showing results from the model trained on 100\% real data, middle column showing results from the best model trained on a mix of real and simulated data (A + S) and right most column showing results from the best model trained on a mix of real and sim2real data (A + G). Fig.~\ref{fig:lane_cross_qualitative} shows qualitative results of the cross-dataset generalization experiments from the paper for the task of lane detection.
The baseline model trained on 100\% real CULane \cite{pan2018spatial} data (first column in Fig.~\ref{fig:lane_cross_qualitative}) results in lots of false negatives (highlighted in red squares) and false positives (highlighted in yellow squares) when tested on the TuSimple\footnote{\url{https://github.com/TuSimple/tusimple-benchmark/tree/master/doc/lane_detection}} dataset. Among the models trained with a mix of real and simulated data from Unreal Engine (A + S), the best model is the one trained on 70\% real and 30\% sim data (second column in Fig.~\ref{fig:lane_cross_qualitative}) and results in fewer false negatives as compared to the baseline, but the number of false positives goes up. Overall, the best results are obtained with a model trained on 40\% real and 60\% sim2real data (A + G) with a significantly reduced number of false positives and negatives (last column in Fig.~\ref{fig:lane_cross_qualitative}).
Fig.~\ref{fig:rgb2depth_cross_qualitative} shows qualitative results of the cross-data generalization experiments for the single-image depth task. The best result is achieved by the model trained on a mix of 40\% real and 60\% sim data (A + S). For sim2real data augmentation, quantitatively, the best result is achieved by the model trained on a mix of 50\% real and 50\% sim2real data (A + G). However, for the sake of fair comparison, results in Fig.~\ref{fig:rgb2depth_cross_qualitative} are shown with the A + G model with 60\% synthetic data. As highlighted by the zoomed-in section within each depth map, it can be clearly seen that training on a mix of real and simulated data improves the quality of depth map, especially around periphery of the vehicle silhouettes (Row 3). Moreover, adding sim2real data to the real dataset improves the quality of the predicted depth maps even further (Row 4).

\begin{figure}[h!]
\centering
\includegraphics[width=.6\linewidth]{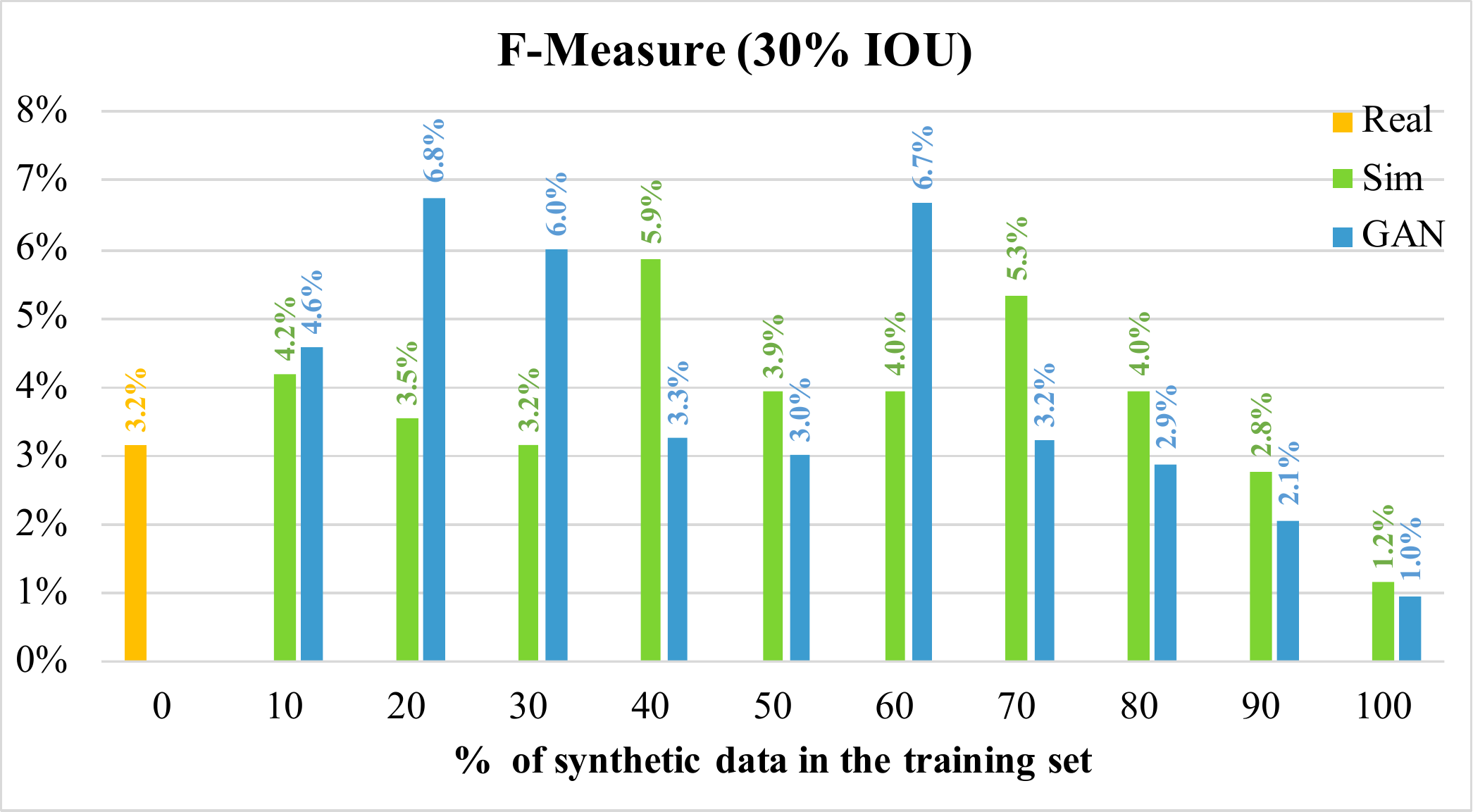}
\caption{Plot of F-measure for cross-dataset testing of lane detection models trained on a mix of real TuSimple (dataset B) and synthetic images (either simulated or sim2real translated) and tested on real CULane (dataset A) images. As you move from left to right, the ratio of synthetic data in the training set increases.\label{fig:fmeasure_cross_test_2}}
\end{figure}

\begin{center}
\begin{figure}[h!]
\centering
    \includegraphics[width=.6\linewidth]{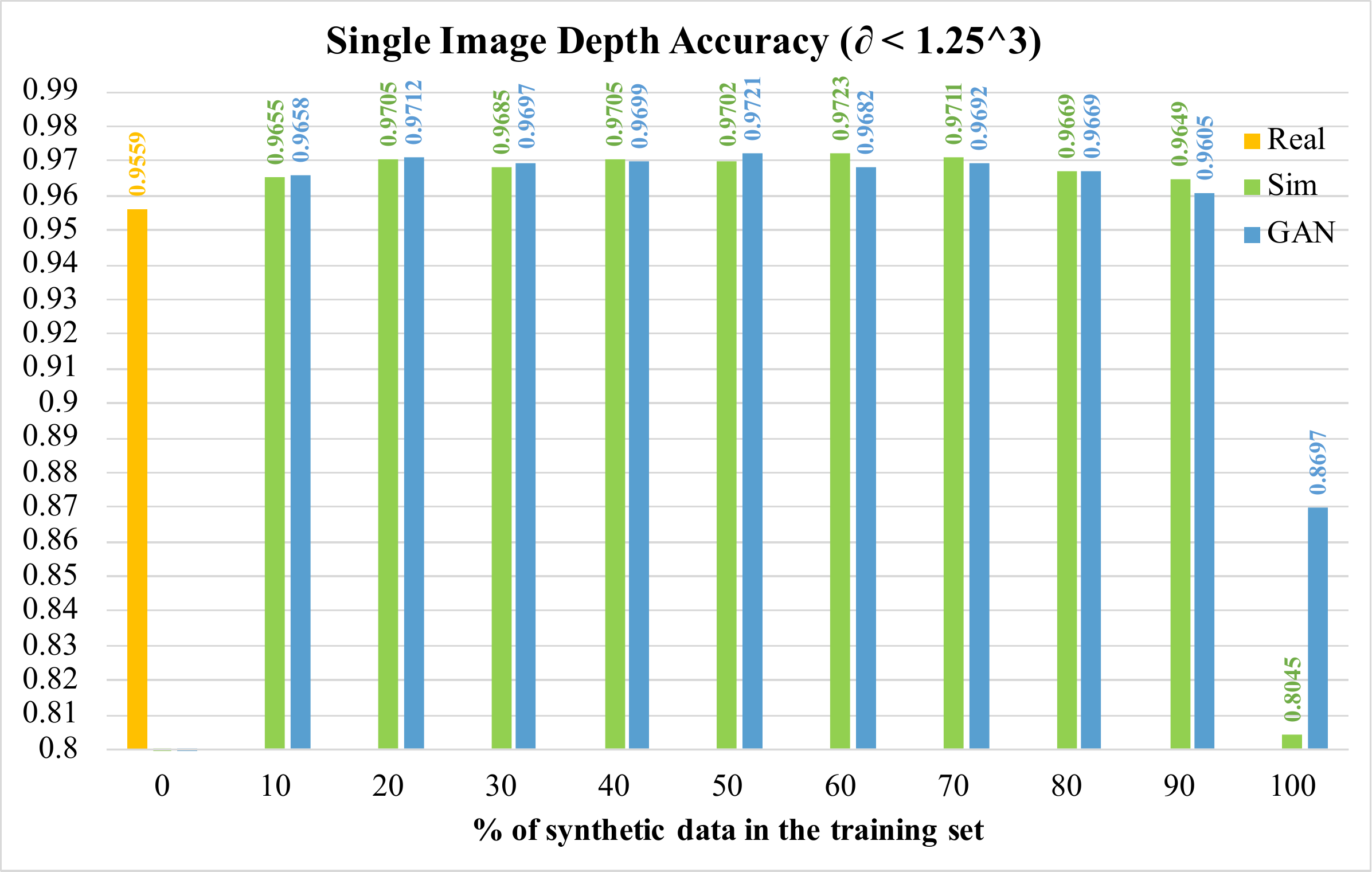}
   \caption{Accuracy results for the single image depth task (higher is better).}
   \label{fig:monodepth_metrics_acc}
\end{figure}
\end{center}

Section~\ref{sec:supp_quant} provides additional quantitative results for the three tasks. In particular, Table.~\ref{tab:experiments_slot_tpfp} provides additional quantitative insights into the role of synthetic data augmentation in improving the number of true positives and false positives for the slot detection task. Table~\ref{tab:experiments_slot_dp} shows how increasing dropout regularization does not help improve generalization performance of models trained on 100\% real data. Fig.~\ref{fig:fmeasure_cross_test_2} and Table~\ref{tab:experiments_lane_b2a} together provide a summary of cross-dataset testing results for lane detection models trained on TuSimple and tested on CULane. Consistent with the results in the main paper from models trained on CULane and tested on TuSimple, synthetic data augmentation helps deflate inherent bias in the TuSimple dataset and improve cross dataset generalization performance. Fig.~\ref{fig:monodepth_metrics_acc} provides additional quantitative results from the cross-dataset generalization experiments for the task of depth estimation. 

\clearpage
\subsection{Noise Factor Distribution of Simulated Data}
\label{sec:noise_factor}
The simulated data for both the parking slot detection and lane detection tasks was generated using an in-house Unreal Engine-based pipeline. Table~\ref{tab:noise_factors_lane} lists the noise factors along with their range of variation that were used to generate the simulated data for the lane detection task. For each noise factor, values were randomly sampled from a uniform distribution within the specified range. The road spline was randomly generated as well but with some checks to ensure the spline was smooth and did not break or loop back on itself. A total of 300 scenarios were thus created with 300 frames in each leading to a total of 90000 simulated images. Table~\ref{tab:noise_factors_slot} lists the noise factors varied for the parking slot detection task to generate 7 simulated scenarios resulting in a total of 15565 images for training.

\begin{table}[h!]
\footnotesize
  \centering
  \caption{List of noise factors along with their range of variation used for generating the simulated data for the lane detection task. All factors except for sun intensity and cloud density are integer values. Factors for which no units are specified are unitless by design.\label{tab:noise_factors_lane}}
  \begin{tabular}{|c||c|c|c|c|c|c|c|}
  \toprule
  \textbf{Noise Factor} & Sun Intensity & Cloud Density & Sun Angle & Traffic Density & Traffic Speed & No. of Lanes & Speed Limit\\
   & & & Pitch + Yaw (deg) & & Std. Dev. & & (mph)\\
  \midrule
  \textbf{Range} & [0, 3] & [0, 2.5] & [0, 180] & [5,20] & [0, 30] & [1, 4] & [50, 90]\\
  \bottomrule
  \end{tabular}%
\end{table}%

\begin{table}[h!]
\footnotesize
  \centering
  \caption{List of noise factors varied for generating simulated data for the slot detection task. The header S.No. stands for scenarios numbers, which indicate the 7 different scenarios simulated based on the noise factors descriptions listed against them.\label{tab:noise_factors_slot}}
  \begin{tabular}{|c|c|c|c|c|c|c|c|c|c|c|c|c|}
  \toprule
  \textbf{S.No.} & \textbf{Weather} & \textbf{Parking} & \textbf{Line} & \textbf{Line} & \textbf{Line} & \textbf{Time} & \textbf{Sun} & \textbf{Sun } & \textbf{Ground} & \textbf{True} & \textbf{Cloud} & \textbf{No. of}\\
   & & \textbf{Density} & \textbf{Color} & \textbf{Damage} & \textbf{Thickness} & \textbf{of Day} & \textbf{Angle} & \textbf{Intensity} & \textbf{Material} & \textbf{Negatives} & \textbf{Opacity} & \textbf{Frames}\\
  \midrule
  1 & clear & heavy & yellow & 1 & 0.05 & 10 am & (-60, 45) & 5 & cracked & trees, signs & 1 & 2084 \\
  2 & clear & medium & white & 0.1 & 0.12 & 8 am & (-30, 45) & 3 & asphalt & trees, signs & 1 & 2071 \\
  3 & clear & medium & yellow & 0.1 & 0.15 & 10 am & (-60, 90) & 10 & cracked & side walk & 1 & 2427 \\
  4 & clear & light & white & 0.5 & 0.15 & 12 pm & (-90, 90) & 8 & cracked & side walk & 3 & 2415 \\
  5 & clear & light & yellow & 0.2 & 0.1 & 2 pm & (-120, 60) & 10 & asphalt & side walk & 1 & 2133 \\
  6 & overcast & heavy & white & 0.5 & 0.1 & 4 pm & (-160, 75) & 1 & cracked & side walk & 3 & 1632 \\
  7 & overcast & light & yellow & 0.5 & 0.05 & 8 am & (-30, 60) & 1 & asphalt & grass & 1 & 2803 \\
  \bottomrule
  \end{tabular}%
\end{table}%

\clearpage
\subsection{Parking Slot Detection: Qualitative Results}
\label{sec:qual_slot}

\clearpage
\subsection{Traffic Lane Detection: Qualitative Results}
\label{sec:qual_lane}

\clearpage
\subsection{Depth Estimation: Qualitative Results}
\label{sec:qual_depth}

\clearpage
\subsection{Case Studies: Additional Quantitative Results}
\label{sec:supp_quant}
\begin{table}[h!]
\footnotesize
  \centering
  \caption{Summary of MobileNetV2 SSD based parking slot detection cross-dataset testing results (train on Parking A, test on Parking B) for varying dropout percentages. F-Measure was 0\% for all models.}
    \begin{tabular}{|c||c|c|c|c|c|}
    \toprule
    \textbf{Dropout} & 0\% & 0.5\% & 1\% & 10\% & 90\%\\
    \midrule
    \textbf{False Positives} ($\downarrow$) & 252 & 58 & \textcolor{green}{\textbf{3}} & 91 & 920\\
    \bottomrule
    \end{tabular}%
  \label{tab:experiments_slot_dp}%
\end{table}%

\begin{table}[h!]
\footnotesize
  \centering
  \caption{Summary of cross-dataset testing results for parking slot detection. Here, TP and FP denote the number of true positives and false positives respectively.}
    \begin{tabular}{|l|l|l|l|}
    \toprule
    \textbf{Train} & \textbf{Test} & \textbf{TP} ($\uparrow$)& \textbf{FP} ($\downarrow$)\\
    \midrule
    A & B & 0 & 252 \\
    $\text{A}+\text{S}\ (40\%)$ & B & 89 & \textbf{35} \\
    $\text{A}+\text{G}\ (50\%)$ & B & \textbf{303} & 149 \\
    \bottomrule
    \end{tabular}%
  \label{tab:experiments_slot_tpfp}%
\end{table}%

\begin{figure}[h!]
\centering
\includegraphics[width=.75\linewidth]{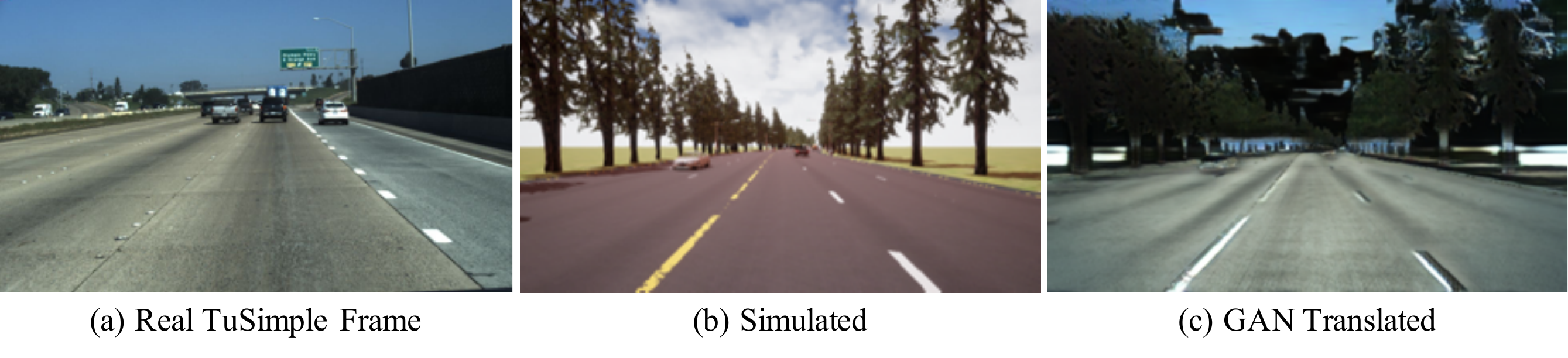}
\caption{Example real, simulated and sim2real translated images used for training models in Fig.~\ref{fig:fmeasure_cross_test_2} and Table.~\ref{tab:experiments_lane_b2a}\label{fig:real_sim_gan_lane_tusimple}}
\end{figure}

\begin{table}[h!]
\footnotesize
  \centering
  \caption{Summary of results in Fig.~\ref{fig:fmeasure_cross_test_2}. Here, A and B denote the CULane and TuSimple datasets respectively. S denotes simulated images and G denotes the sim-to-real GAN translated equivalent of S. Example B, G and D images are shown in Fig.~\ref{fig:real_sim_gan_lane_tusimple}. For all these experiments, SCNN was trained on $512\times288$ images. For cross-dataset testing, CULane images were downsized and then padded (along height) to match the training resolution of $512\times288$ while simultaneously maintaining the original aspect ratio. For synthetic data augmentation rows, results are shown for the best model in terms of F-measure on cross-dataset testing in green for A + S and in blue for A + G.}
    \begin{tabular}{|c|c|c|c|c|}
    \toprule
    \textbf{Train} & \textbf{Test} & \textbf{Precision} ($\uparrow$)& \textbf{Recall} ($\uparrow$)& \textbf{F-Measure} ($\uparrow$)\\
    \midrule
    B & B & 80.2\% & 91.7\% & 85.6\%\\
    $\text{B}+\text{S}\ (40\%)$ & B & 80.1\% & 91.5\% & 85.4\% \\
    $\text{B}+\text{G}\ (20\%)$ & B & 79.1\% & 90.1\% & 84.2\% \\
    \midrule
    B & A & 2.8\% & 3.7\% & 3.2\%\\
    $\text{B}+\text{S}\ (40\%)$ & A & \textcolor{green}{\textbf{5.3\%}} & \textcolor{green}{\textbf{6.6\%}} & \textcolor{green}{\textbf{5.9\%}} \\
    $\text{B}+\text{G}\ (20\%)$ & A & \textcolor{blue}{\textbf{5.9\%}} & \textcolor{blue}{\textbf{7.8\%}} & \textcolor{blue}{\textbf{6.8\%}} \\
    \bottomrule
    \end{tabular}
    \label{tab:experiments_lane_b2a}
\end{table}%

\end{document}